\documentclass{article}

\usepackage{jmlr2e}

\usepackage[nottoc]{tocbibind}
\usepackage{parskip}    %

\usepackage[round]{natbib}
\setcitestyle{authoryear,round,citesep={;},aysep={,},yysep={;}}

\usepackage[T1]{fontenc}    %
\usepackage{url}            %
\usepackage{booktabs}       %
\usepackage{amsfonts}       %
\usepackage{nicefrac}       %
\usepackage{microtype}      %
\usepackage{graphicx}
\usepackage[dvipsnames]{xcolor}
\colorlet{Blue}{Blue}
\usepackage{todonotes}
\RequirePackage{doi}
\usepackage{amsmath}
\usepackage{amsthm}

\newtheorem{theorem}{Theorem}
\newtheorem{corollary}{Corollary}[theorem]

\usepackage{bm}
\usepackage{subcaption}
\usepackage{enumitem}
\usepackage{wrapfig}
\usepackage{mathtools}
\usepackage{pgffor}
\usepackage{grffile}
\usepackage{tikz}
\usepackage[export]{adjustbox}

\usetikzlibrary{positioning}
\usetikzlibrary{arrows.meta,decorations.pathreplacing, decorations.markings}
\tikzstyle{vecArrow} = [thin, decoration={markings,mark=at position
   1 with {\arrow[<]{open triangle 30}}},
   double distance=1.4pt, shorten >= 5.5pt,
   preaction = {decorate},
   postaction = {draw,line width=1.4pt, white,shorten >= 4.5pt}]
\tikzstyle{innerWhite} = [semithick, white,line width=1.4pt, shorten >= 4.5pt]
\usetikzlibrary{patterns}
\usetikzlibrary{bayesnet}
\tikzstyle{disent_latent} = [circle,pattern=north east lines, pattern color=black!20,draw=black,inner sep=1pt,
minimum size=20pt, font=\fontsize{10}{10}\selectfont, node distance=1]
\usepackage{multirow}
\usepackage{subfiles}
\usepackage{varioref}
\usepackage{etoolbox}
\usepackage{IEEEtrantools}
\usepackage{xr}
\usepackage{xr-hyper}
\usepackage{hyperref} 
\makeatletter

\newcommand*{\addFileDependency}[1]{%
  \typeout{(#1)}
  \@addtofilelist{#1}
  \IfFileExists{#1}{}{\typeout{No file #1.}}
}
\makeatother

\newcommand*{\myexternaldocument}[1]{%
    \externaldocument{#1}%
    \addFileDependency{#1.tex}%
    \addFileDependency{#1.aux}%
}
\myexternaldocument{supp}

\DeclareMathOperator*{\argmin}{argmin}

\DeclareMathOperator{\expect}{\mathbb{E}}
\DeclareMathOperator*{\KL}{KL}

\DeclareMathOperator*{\ent}{\mathcal{H}}
\DeclareMathOperator{\ELBO}{\mathcal{L}}

\DeclareMathOperator{\I}{\mathbb{I}}
\DeclareMathOperator{\cat}{\mathrm{Cat}}

\renewcommand{\vector}[1]{\boldsymbol{\mathbf{#1}}}
\renewcommand{\v}{\vector}
\newcommand*\vv[1]{\vec{\v{#1}}}

\makeatletter
\newif\ifnobrackets
\renewcommand\@cite[2]{\ifnobrackets\else[\fi{#1\if@tempswa , #2\fi}\ifnobrackets\else]\fi\nobracketsfalse}
\newcommand\nbcite{\nobracketstrue\cite}
\makeatother

\def\arrvline{\hfil\kern\arraycolsep\vline\kern-\arraycolsep\hfilneg}

\title{Relaxed-Responsibility Hierarchical Discrete VAEs}

\author{\name{Matthew Willetts$^{1,2}$} \email{mwilletts@turing.ac.uk}\\
\name{Xenia Miscouridou$^{1,2}$} \email{xmiscouridou@turing.ac.uk}\\
\name{Stephen Roberts$^{2,3}$} \email{sjrob@robots.ox.ac.uk}\\
\name{Chris Holmes$^{1,2}$} \email{cholmes@stats.ox.ac.uk}\vspace{1em}\\
\addr ${}^{1}$Department of Statistics, University of Oxford\\
${}^{2}$Alan Turing Institute, London\\
${}^{3}$Oxford-Man Institute, University of Oxford
\vspace{-4em}
}

\begin{document}

\maketitle

\begin{abstract}
Successfully training Variational Autoencoders (VAEs) with a hierarchy of discrete latent variables remains an area of active research.
 Vector-Quantised VAEs are a powerful approach to discrete VAEs, but naive hierarchical extensions can be unstable when training.
Leveraging insights from classical methods of inference we introduce \textit{Relaxed-Responsibility Vector-Quantisation}, a novel way to parameterise discrete latent variables, a refinement of relaxed Vector-Quantisation that gives better performance and more stable training.
This enables a novel approach to hierarchical discrete variational autoencoders with numerous layers of latent variables (here up to 32) that we train end-to-end.
Within hierarchical probabilistic deep generative models with discrete latent variables trained end-to-end, we achieve state-of-the-art bits-per-dim results for various standard datasets.
Further, we observe different layers of our model become associated with different aspects of the data.

\end{abstract}
\section{Introduction}
\label{sec:intro}
Probabilistic deep generative models, such as Variational Autoencoders (VAEs), have had significant and continuing success in learning continuous representations of data~\citep{Kingma2013, Rezende2014, Kingma, Vahdat2020, Child2020}.
The learning of discrete representations has also flourished \citep{Grathwohl2017, VQVAE, Razavi2019, Fortuin2019, Pervez2020} and remains an active area of research.
Discrete representations are useful as they are intrinsically compact, finding application in various tasks such as compression and clustering.
Advances in differentiable relaxations of discrete probability distributions~\citep{Maddison2016,Jang2016} have contributed to training discrete latent variables models on high-dimensional data using gradient-based methods~\citep{Sonderby2017}.
However, training rich hierarchical models with discrete latent variables for high-dimensional data remains a problem in the field~\citep{Lievin2019, Williams2020,Pervez2020}.

Here we propose an effective, scalable method for learning hierarchical discrete representations of image data within a unified probabilistic framework.
This work builds on Vector-Quantised Variational Autoencoders (VQ--VAEs)~\citep{VQVAE} and their relaxation~\citep{Sonderby2017}.

\begin{figure*}[t!]
\centering
\begin{minipage}[c]{\textwidth}
\begin{subfigure}[c]{\textwidth}
\begin{minipage}[c]{\textwidth}
\includegraphics[width=\textwidth]{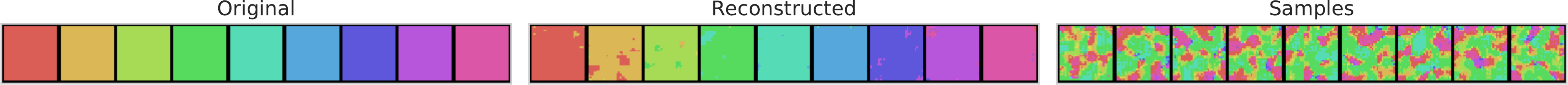}\\[-1.2pt]
\includegraphics[width=\textwidth]{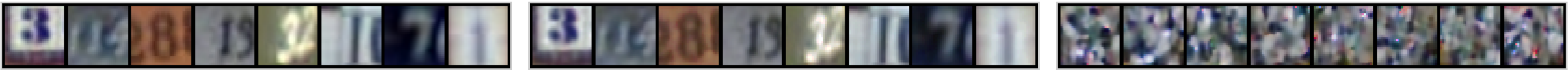}\\[-1.3pt]
\includegraphics[width=\textwidth]{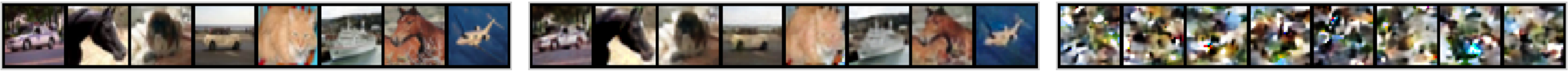}\end{minipage}
\caption{rVQ--VAE}\label{fig:Rel_example}
\end{subfigure}
\\[1ex]
\begin{subfigure}[c]{\textwidth}
\begin{minipage}[c]{\textwidth}
\includegraphics[width=\textwidth]{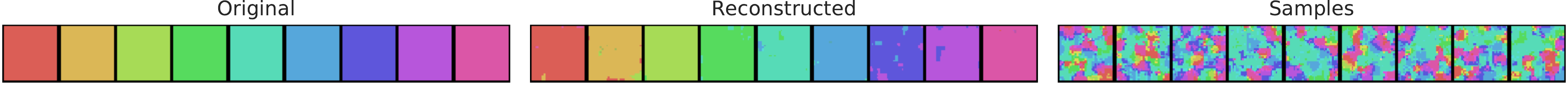}\\[-1pt]
\includegraphics[width=\textwidth]{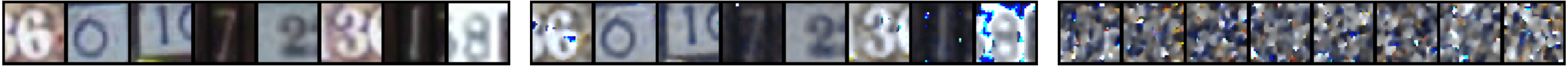}\\[-1pt]
\includegraphics[width=\textwidth]{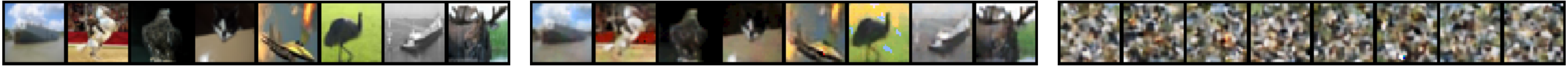}\end{minipage}
\caption{Spatial--VAE}\label{fig:Cont_example}
    \end{subfigure}%
    \caption{Here we demonstrate that the poor quality draws when sampling from a VQ--VAE's prior $p(\v{z})$ is not from having discrete latents, but from the spatial arrangement of latent variables. We train
    (a) rVQ--VAEs and (b) Spatial--VAEs (a VAE with continuous latents, but arranged spatially like a VQ--VAE) on (top) a toy dataset composed of 9 colour swatches, (middle) SVHN, (bottom) CIFAR-10. For each dataset, both models give good reconstructions (middle column) but ancestral samples from the prior $p(\v{z})$ (right column) are very dissimilar to datapoints in the training set, even for the toy dataset -- for which we do not see uniformly-coloured images, instead we see regions of each the different colours of the dataset. This shows that it is the method used to parameterise the model's latent variables that leads to this sampling phenomena, not being discrete vs continuous.}
        \label{fig:colour}
    \end{minipage}
\end{figure*}

VQ--VAEs reach surprisingly poor raw bits-per-dim (bpd), a scaled form of the ELBO, during training, on both train and test set and thus to achieve good performance they require post-hoc training of density estimators on learnt embeddings.
We begin by analysing how this happens.
Perhaps one might think it is because of the probabilistic structure of these models -- that having discrete latent variables as the prior leads to poor generations.
We find that VAEs with the same neural parameterisation as VQ--VAEs -- convolutional neural networks with latents laid out spatially -- but with Gaussian latent variables show the same pathologies in training.
See Fig~\ref{fig:colour} for a demonstration of this.

This motivates us to develop a novel variety of hierarchical discrete VAEs.
Previously developed hierarchical structures based around VQ building blocks have required various heuristics in model formulation and training~\citep{Williams2020} or highly restricted probabilistic structure~\citep{Pervez2020}.
We find that naive hierarchical extensions can be unstable during training.
With a new formulation of probabilistic vector quantisation we train hierarchical discrete latent variable models end-to-end within a unified probabilistic framework.
These models, which we call \textit{Relaxed-Responsibility Vector-Quantised VAEs} or RRVQ--VAEs, have a hierarchical structure that means they achieve state of the art bits-per-dim for this class of models.

Our models show superior performance when compared against VQ--VAE baselines and naive hierarchical extensions, as well as various baselines.
We find that performance increases as we increase the number of layers of latent variables in our model with the deepest models we train having 32 layers.
Further, we demonstrate that our model places information about different aspects of the images into different latent layers.
We also demonstrate that our approach can be used to perform compression.

RRVQ-VAEs help to close the performance gap between discrete VAEs and their continuous counterparts.
This approach opens up new avenues for the building of hierarchical discrete VAEs and is a step towards a unified probabilistic framework for specifying and training models of this type.

\section{Background: Vector Quantised Variational Autoencoders}

The Vector-Quantised Variational Autoencoder (VQ--VAE)~\citep{VQVAE} is a density estimator for high dimensional data such as audio, images and video.
Instead of having continuous latent variables, as in the vanilla VAE (see Appendix~\ref{app:A}), the latents $\v{z}$ are a set of $M$ discrete variables $\v z=\{z^1,\dots,z^M \}$ each of dimensionality $K$.
The joint $p_\theta(\v{x}, \v{z})$ factorises as for a vanilla VAE, but with $p(\v{z})=\prod_{m=1}^{M} \cat\left(z^m|\frac{1}{K}\right)$.

The likelihood $p_\theta(\v{x}|\v{z})$ does not depend directly on samples of $\v{z}$.
Rather the discrete vector $\v{z}$ is used to index over a dictionary of $K$ embeddings, the codebook vectors $\v{E}=\{\v{E}^k\}$, each $\v{E}^k\in\mathbb{R}^{d_e}$, $d_e$ being the dimensionality of the embedding space. %
For stochastic amortised variational inference in VQ--VAEs, introduce a recognition network $\v{e}_\phi(\v{x}) \in \mathbb{R}^{M\times d_e}$ outputing $M$ vectors in $\mathbb{R}^{d_e}$, the embedding space.
The posterior $q_\phi(\v{z}|\v{x}) = \prod_{m=1}^M q_\phi(z^m|\v{x})$ is then defined via a nearest-neighbour vector-lookup.
For each latent $z^m$,
\begin{align}
    q_\phi\left(z^m = k|\v{x}\right) & = \left\{ \,
\begin{IEEEeqnarraybox}[][c]{l?s}
\IEEEstrut
1 & if $k$ = $\argmin_j{\left|\v{e}_\phi^m\left(\v{x}\right) - \v{E}^j \right|^2_2}$ \\
0 & otherwise.
\IEEEstrut
\end{IEEEeqnarraybox}
\right.
\end{align}
This is a one-hot posterior: $q_\phi(\v{z}|\v{x})$ is deterministic.
In a vanilla VAE we train the model by maximising the ELBO, $\ELBO(\v{x})=\expect_{\v{z}\sim q}\log p_\theta\left(\v{x}|\v{z}\right) - \KL\left(q_\phi\left(\v{z}|\v{x}\right)||p(\v{z})\right)$, over the dataset with respect to the generative and recognition model parameters.
This standard training approach is not appropriate for this discrete model, for two reasons.
Firstly, since it is not possible to differentiate through the vector lookup operation (due to the $\mathrm{argmin}$) we cannot use differentiable samples to take gradients through Monte Carlo estimates of the expectations.
Secondly, the one-hot posterior makes the $\KL$ term constant (equal to $M\log K$) so there is no regularisation on the posterior representations.

Thus, a VQ--VAE has two extra terms in its objective: a vector quantisation loss to train the embeddings; and a commitment loss to control the output of the embedding network, weighted by a chosen hyperparameter $\beta$~\citep{VQVAE}.

\paragraph{rVQ--VAEs}
Instead of the deterministic posterior found in a vanilla VQ--VAE, a Gumbel-Softmax distribution~\citep{Maddison2016,Jang2016} can be used to specify a posterior distribution from which we can take differentiable samples~\citep{Sonderby2017}.
This means that the posterior $q_\phi(\v{z}|\v{x})$ is no longer a one-hot, deterministic distribution and the $\KL$ in $\ELBO$ is no longer a fixed constant.
Thus the VQ-related loss terms are no longer needed and the codebook can be learnt via gradient descent.
One can choose the logits of the posteriors to be proportional to the square distance between the given embedding vector and each codebook vector~\citep{Sonderby2017},
\begin{align}
 q(\v{z}|\v{x})&=\prod_{m=1}^M\cat\left(\v{z}_{m}|\pi_\phi^{m}\left(\v{x}\right)\right) \label{eq:gssoft_1}
\end{align}
\begin{align}
\pi_\phi^{m,k}(\v{x})&\propto\exp\left(-\frac{1}{2}\left|\v{e}_\phi^m(\v{x})-\v{E}^k\right|_2^2\right) \label{eq:gssoft_2}.
\end{align}
These \textit{Relaxed}--VQ--VAEs (henceforth rVQ--VAEs) have been shown to make better use of their latent variables than the deterministic base model, obtaining higher values of $\ELBO$~\citep{Sonderby2017} both at train and test time.

Both Vanilla--VQ--VAEs and rVQ--VAEs train relatively stably.
By definition Vanilla--VQ--VAEs avoid posterior collapse \citep{Bowman2016, Razavi2019a, Dai2019, Lucas2019} as the $\KL$ term in $\ELBO$ is constant.
For rVQ--VAEs matching the posterior to the prior in the latent space is not possible in the general case as it would require the posterior embedding to be equidistant from all codebook vectors.

In this work we build on rVQ--VAEs, not deterministic VQ--VAEs, both due to their demonstrated superior performance in maximising $\ELBO$ and as they truly have probabilistic structure.
In the rest of this paper, `VQ--VAE' is used both to refer generically to either Vanilla (ie, deterministic) VQ--VAEs and to rVQ--VAEs, as their properties and behaviours are broadly similar.
When needing to refer to them distinctly, we do so.

\section{Sampling and Reconstructing in VQ--VAEs}
\label{sec:sampling}
Here we focus on modelling square images, though the arguments we make can generalise to images, as well as to audio or video data.
In VQ--VAEs, one uses convolutional neural networks to represent $p$ and $q$, laying out $\v{z}$ as a square of side $\sqrt{M}$, mirroring the spatial structure of pixels in an image \citep{VQVAE}.
For audio one might choose a 1D structure, and 3D for video.

Interestingly, ancestral sampling from $p_\theta(\v{z})$ in (relaxed or not) VQ--VAE models gives draws that do not resemble the training data.
This indicates severe aggregate posterior--prior mismatch.
Samples from this prior fail to capture the structure needed, i.e. the dependencies between the $M$ latents that are necessary to produce realistic data when decoded.

Meanwhile, even from early stages of training in VQ--VAEs the reconstructions of training data are of high fidelity.
This is why in VQ--VAEs it is necessary to subsequently train a second density estimator, commonly a large, powerful autoregressive model such as a PixelCNN~\citep{VandenOord2016a,Salimans2017} over the latent representations to then sample from.
This is followed in the two- and three-layer extension of VQ--VAEs as well~\citep{Razavi2019}.

Conversely, in VAEs with continuous latent variables the reconstructions are generally found to be somewhat blurry, while samples tend to have more coherent structure.
In a standard VAE with $p(\v{z})=\prod_{i=1}^M \mathcal{N}\left(z^m|0,1\right)$ the prior factorises over dimensions similar to how it does in a VQ--VAE, yet samples appear reasonable, which suggests that the reason is not only that.

We give an explanation for this phenomenon.
It is not related with discrete vs continuous latents at all, but rather with their neural parameterisation:
In VQ--VAEs, convolutional neural networks are used to represent $p$ and $q$.
With convolutionally-parameterised latents, each is tied spatially to be mostly concerned with a particular region of pixels in the input.
This is unlike most implementations of vanilla VAEs, where the posterior's parameters, commonly the mean and diagonal covariance of a Gaussian, and the decoder mean are output by MLPs.
Those learnt representations are thus intrinsically non-local,
which in turn gives them the ability to learn easily the arrangement of parts and wholes in an image.

To demonstrate this, we train a simple \textit{Spatial}--VAE where continuous-valued latent variables are arranged spatially, as in VQ--VAEs: $p_\theta(\v{z})=\prod_{m=1}^M\mathcal{N}\left(\v{z}^m|\v{0},\mathbb{I}\right)$ and $q_\phi(\v{z}|\v{x})=\prod_{m=1}^M\mathcal{N}(z^m|\mu^m_\phi(\v{x}),\sigma^m_\phi(\v{x}))$, $\v{z}^m\in\mathbb{R}^{16}$, with $p$ and $q$ convolutional networks each composed of 2 ResNet block with 32 channels, and the number of latents $M$ is the $1/4$ the number of pixels in the input.
We also train an equivalent rVQ--VAE, with embedding space dimensionality $d_e=K=16$.
We use SVHN, CIFAR-10, and (to make the effect most striking) a toy dataset containing images that are each uniform blocks of colours.
See Fig~\ref{fig:colour} for the resulting reconstructions and samples for the three datasets for both models.
We also provide examples of toy MLP-parameterised VQ-VAEs providing coherent samples in Appendix~\ref{app:mlp}.

Embedding an image into the latent space for reconstruction is relatively easy.
For the discrete model, with high probability, the encoder outputs embeddings $\v{e}_\phi(\v{x})$ that are close to the appropriate codebook embedding, (appropriate given its local region of the image) and this is the case for each of the $M$ spatially-arranged latents.
Similarly, at each latent position the Spatial--VAE encoder learns to place posterior probability over the appropriate latent space region.
However, when sampling from each model's prior, we end up with very mixed up generated images.
Even for the toy dataset, the draws for both models are rainbow images where each patch of the image is separately given a random colour from the training set.

The poor quality of naive VQ--VAE draws is not intrinsically from having discrete latent variables, but from having discrete latent variables \textit{that are arranged spatially having been parameterised using convolutional neural networks}.
However, it is the choice to have spatial latent variables that provides high quality reconstructions.
A solution around this is to train a powerful autoregressive model over samples from the aggregate posterior in $\v{z}$.
In Vanilla--VQ--VAEs the aggregate posterior is a sum of $\delta$ functions, so it resembles an empirical data distribution.
Thus training a high-performance density estimator is reasonable and provides realistic draws~\citep{VQVAE, Razavi2019}.
In this manner of operation, the encoder-decoder networks can be viewed as tools for non-linear dimensionality-reduction, so that the density estimator can be trained in a lower-dimensional space, the learnt latent space, rather than on the raw data directly.
While that is a proven and performative approach, our goal is to combine the benefits of VQ--VAEs (high quality reconstructions, the desirable property of learning discrete representations, ease of training) with having a unified modelling approach, with models trained end-to-end.

We develop ways to make discrete VAEs more expressive and flexible by adding hierarchical structure.
This removes the need of a two-stage training process, and gives us the benefits of hierarchical representations such as having different layers learning different aspects of the data. Further, having autoregressive models for sampling, when using the trained model would require to perform as many forward passes through the model as there are latent variables. In the hierarchical case, as in \cite{Razavi2019}, this is still true.

In this paper we will be training very deep hierarchies of latent variables, up to 32 layers.
Therefore, if we had autoregressive models for sampling, the additional calls that would be needed to produce one sample from a hierarchical model of this type, would be a very difficult requirement.
For our deepest models trained on $32\times32$ images it would be $\approx 2000$ internal, sequential forward passes.
For $64\times 64$ images it would be $\approx 10,000$.
Instead, with our approach, we are able to generate samples using a single forward pass.

\section{Relaxed-Responsibility Hierarchical Discrete VAEs}
\subsection{Hierarchical Discrete VAEs}
To make a hierarchical discrete VAE, introduce $L$ layers of latent variables $\vec{\v{z}}=\{\v{z}_1,..,\v{z}_L\}$. Note that $\v{z}_\ell^m$ is the $m^{\mathrm{th}}$ latent variable in the $\ell^{\mathrm{th}}$ layer.
We wish to have an autoregressive structure between layers.
Inspired by the ResNet VAEs~\citep{Kingma}, we choose our generative model's factorisation to be
\begin{align}
p_\theta\left( \v {x},\vec{\v{z}}\right) = p_\theta\left(\v{x}|\vec{\v{z}}\right)p_\theta \left(\vec{\v{z}}\right) = p_\theta(\v{x}|\vec{\v{z}})p(\v{z}_L)\prod_{\ell=1}^{L-1} p\left(\v{z}_\ell|\v{z}_{>\ell}\right)
\label{eq:p_lad}
\end{align}
where $p(\v{z}_\ell|\v{z}_{>\ell})=\cat\left(\v{z}_\ell|f^\theta_\ell(\v{z}_{>\ell})\right)$ and $p_\theta(\v{z}_L)=\cat\left(\v{z}_L|\pi_\theta^L\right)$.
Similarly, $q$ factorises as
\begin{align}
q_\phi(\vv{z}|\v{x})=q_\phi(\v{z}_L|\v{x})\prod_{\ell=1}^{L-1}q_\phi(\v{z}_\ell|\v{z}_{>\ell},\v{x}).
\label{eq:q_lad}
\end{align}
The ELBO $\mathcal{L}(\v{x})$ for this model is thus
\begin{align}
    \mathcal{L}(\v{x}) =&\expect_{\vv{z}\sim q} \log p_\theta(\v{x}|\vv{z}) - \KL\left(q_\phi(\v{z}_L|\v{x})||p(\v{z}_L\right) \nonumber \\
    &-\sum_{\ell=1}^{L-1} \expect_{\v{z}_{>\ell}\sim q}\KL(q_\phi(\v{z}_\ell|\v{z}_{>\ell}, \v{x})||p_\theta(\v{z}_\ell|\v{z}_{>\ell})  \label{eq:lvae_elbo}.    
\end{align}
This is directly analogous to hierarchical VAEs with continuous latent variables.

\subsection{Relaxed-Responsibility Vector-Quantisation}
\label{sec:rrvq}
Our first main contribution is a method of parameterising the generative model and the approximate posterior for models containing vector-quantised discrete latents. This improves the ability of hierarchical models of this type to learn effectively. We call this method \textit{Relaxed-Responsibility Vector-Quantisation} (RRVQ).
We found that without these improvements, models of this form had low performance and were often unstable during training, and that the two changes we propose are synergistic -- working better together than either alone.

\subsubsection{Proposal for $q$}
Vector-Quantisation has historic links to mixture models, mixtures of experts, and classical methods of inference.
The exponential moving average method of updating the codebook in VQ--VAEs is closely linked to K-means \citep{macqueen1967}.
rVQ is linked to mean-field variational inference for a mixture of Gaussians:
we can interpret the embedding codebook as recording the means of the cluster components, all having isotropic unit variance, and are a-priori equal in probability~\nbcite[\S10.2]{Bishop2006}.
Eq~\eqref{eq:gssoft_2} is equivalent to saying that the posterior at each position in $\v{z}$ is equal to the cluster responsibilities for the embedding vector $\v{e}_\phi(\v{x})$ at that position.

We develop this link further, increasing the expressiveness of the parameterisation of the latents $\v{z}_\ell$ in our hierarchical model, by relaxing the restriction that all components have unit isotropic covariance.
We introduce a second codebook $\v{E}_{\Sigma,\ell}$ for each layer, recording the diagonal covariance matrices of each component.
The responsibilities then used for defining $\pi_{\phi,\ell}(\v{e}_{\phi,\ell})$ are
\begin{equation}
\pi_{\phi,\ell}^{m,k}(\v{e}_{\phi,\ell}) \propto \frac{\exp{\left(-\frac{1}{2\v{E}^k_{\ell,\Sigma}}\left|\v{e}_{\phi,\ell}^m-\v{E}_{\mu,\ell}^k\right|_2^2\right)}}{\sqrt{\left(2\pi\right)^{d_e}\v{E}^k_{\ell,\Sigma}}},
\label{eq:rr_q}
\end{equation}
where $m$ indexes over the latent positions, $k$ over the codebook entries, $\v{e}_{\phi,\ell} \in\mathbb{R}^{M\times d_e}$, $\v{E}_{\mu,\ell}$ is the codebook of means for the $\ell^{\mathrm{th}}$ layer and $\v{e}_{\phi,\ell}$ is the embedding-space output of a network taking the appropriate inputs for the current layer, as written in Eq~\eqref{eq:q_lad}.

Viewing VQ as a mixture-of-experts model~\citep{Jacobs1991}, where each codebook embedding mean is a local expert, we can view this extension as allowing the neighbourhoods of different experts to be more diffuse or more concentrated.
By learning $\v{E}_{\Sigma,\ell}$, codebook embeddings with large diagonal covariance will have their means used preferentially when the output embeddings $\v{e}_{\phi,\ell}$ are far away from the codebook means, and those with small diagonal covariance will dominate at short ranges, being highly confident of being the appropriate expert when $\v{e}_{\phi,\ell}$ is close.

\subsubsection{Proposal for $p$}
One obvious approach is to parameterise the (log) probabilities of $p_\theta(\v{z}_\ell|\v{z}_{>\ell})$ directly by a deep net.
However, we found training to be unstable in hierarchical VQ--VAEs.

Training pathologies in VAEs come from large $\KL$ values.
These large values come from highly-confident distributions that have limited overlap between them, i.e low entropy distributions.
These highly-confident distributions are the result of the underlying neural network outputs taking large-magnitude values.
It is thus reasonable to hope that distributions that have higher-entropy when given large neural network outputs will lead to more stable training.

What is a reasonable, flexible form for the generative model that will provide stable training while preserving or even improving performance?
We might expect rVQ parameterisation of discrete variables to lead to less-peaked, higher-entropy distributions, than a naive implementation using a softmax of logits -- the method we found to be unstable.

To that end, we consider the functional form of the entropy (i) of rVQ-parameterised categorical distributions and (ii) of categorical distributions obtained as the softmax of a vector of logits.
For simplicity, for rVQ  we consider the case $\v{E}_\Sigma = \v 1$, i.e. Eq~\eqref{eq:gssoft_2}, but in \S~\ref{sec:ablate} we show that $p(\v z)$ parameterised either this way or with learnt $\v{E}_\Sigma$ both lead to stable training -- though, as we would predict, learning  $\v{E}_\Sigma$ increases performance.
In Theorems~\ref{th:1} and~\ref{th:2} we consider the worst-case arrangement of codebook means/logits for rVQ and Softmax respectively, such that a single large-magnitude value of the underlying network outputs has maximum impact driving the resulting discrete distribution to be close to one-hot.

\begin{theorem}(Minimum entropy from rVQ)
Consider the worst-case arrangement of rVQ codebooks vectors, i.e. resulting in the minimum entropy categorical distribution: all but one of the codebook embeddings are an equal and greater distance away from the input embedding.
For large-magnitude input embeddings of distance $d$ from the solitary, closest codebook embedding along the line of separation and the remaining $K-1$ codebook embeddings at a distance $d + \delta$ along the same line of separation, the entropy of the resulting categorical distribution, Eq~\eqref{eq:gssoft_2} is, to first order
\begin{equation}
    \mathcal{H}_{\mathrm{rVQ}} \approx \left(K-1\right)\left(1+g\right)\exp{\left(-g\right)},
\end{equation}
where
\(g=\left(\frac{\delta^2}{2}+ \delta d\right)\).
\\Proof: See Appendix~\ref{app:theorem1}.
\label{th:1}
\end{theorem}

\begin{theorem}(Minimum entropy from Softmax)
Consider the worst-case arrangement of logits, i.e. resulting in the minimum entropy categorical distribution: all but one of the logits take the same value $c$, with one logit taking the larger value $c+\ell$, $\ell>0$.
For large-magnitude difference in logits $\ell$, the entropy of the resulting categorical distribution is, to first order,
\begin{equation}
    \mathcal{H}_{\mathrm{softmax}} \approx (K-1)\left(1+\ell\right)\exp{\left(-\ell\right)}
\end{equation}
Proof: See Appendix~\ref{app:theorem2}.
\label{th:2}
\end{theorem}

\begin{corollary}
Viewing $\ell+c=d$ as the large-magnitude output of a neural network, for large d rVQ-parameterised categorical distributions have higher entropy than softmax-parameterised ones if $\delta < 1$.
\end{corollary}
This tells us that for large neural network outputs, rVQ-parameterised distributions have higher entropy than those parameterised via logits, as long as the largest distance between codebooks is $<1$.
We experimentally verify this in Appendix~\ref{app:entropy}.

Thus we choose to parameterise the conditional distributions in $p_\theta(\vv{z})$ along the same lines as for $q$, namely rVQ-parameterisation via an embedding space, rather than using logits, sharing the same codebooks as for $q$.
Given by embeddings $\v{e}_{\theta,\ell}\in\mathbb{R}^{M\times d_e}$ output by a deep net:
\begin{equation}
\pi_{\theta,\ell}^{m,k}\left(\v{e}_{\theta,\ell}\right) \propto \frac{\exp{\left(-\frac{1}{2\v{E}^k_{\ell,\Sigma}}\left|\v{e}_{\theta,\ell}^m-\v{E}_{\mu,\ell}^k\right|_2^2\right)}}{\sqrt{\left(2\pi\right)^{d_e}\v{E}^k_{\ell,\Sigma}}}.
\label{eq:rr_p}
\end{equation}

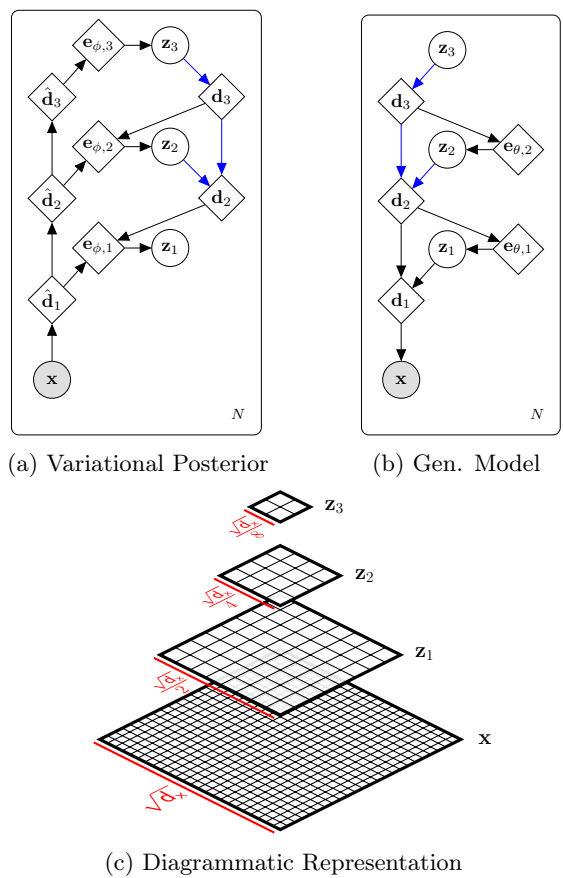
\begin{figure}[t!]
\noindent\makebox[\columnwidth]{%
\subfloat[Variational Posterior]{\scalebox{0.7}{\raisebox{0ex}{
\tikz{ %
      \node[obs] (x) {$\v{x}$} ; %
    \node[det, above=0.7cm of x,minimum size=25pt] (d1) {$\hat{\v{d}}_1$} ; %
    \node[det, above=of d1,minimum size=25pt] (d2) {$\hat{\v{d}}_2$} ; %
    \node[det, above=of d2,minimum size=25pt] (d3) {$\hat{\v{d}}_3$} ; %
    \node[det, above right=0.5cm and 0.4cm of d1] (e1) {$\v{e}_{\phi,1}$} ; %
    \node[det, above right=0.5cm and 0.4cm of d2] (e2) {$\v{e}_{\phi,2}$} ; %
    \node[det, above right=0.5cm and 0.4cm of d3] (e3) {$\v{e}_{\phi,3}$} ; %
    \node[latent, right=0.5cm of e1] (z1) {$\v{z}_1$} ; %
    \node[latent, right=0.5cm of e2] (z2) {$\v{z}_2$} ; %
    \node[latent, right=0.5cm of e3] (z3) {$\v{z}_3$} ; %
    \node[det, below right=0.5cm and 0.5cm of z3,minimum size=25pt] (d_3) {$\v{d}_3$} ; %
    \node[det, below right=0.5cm and 0.5cm of z2,minimum size=25pt] (d_2) {$\v{d}_2$} ; %
	\edge {x} {d1} ; %
	\edge {d1} {d2, e1} ; %
	\edge {d2} {d3, e2} ; %
	\edge {d3} {e3} ; %
	\edge {e1} {z1} ; %
	\edge {e2} {z2} ; %
	\edge {e3} {z3} ; %
	\edge[blue] {z3} {d_3} ; %
	\edge {d_3} {e2} ; %
	\edge[blue] {d_3} {d_2} ; %
	\edge[blue] {z2} {d_2} ; %
	\edge {d_2} {e1} ; %
    \plate[inner sep=0.3cm] {plate1} {(x) (z1) (z2) (z3) (e1) (e2) (d1) (d2) (d3) (d_2) (d_3)} {\scalebox{1}{{$N$}}}; %
  }}
  \label{fig:gout}
  }}
\hspace*{9mm}
\subfloat[Gen. Model]{\scalebox{0.7}{
\tikz{ %
      \node[obs] (x) {$\v{x}$} ; %
    \node[det, above=0.7cm of x,minimum size=25pt] (d1) {$\v{d}_1$} ; %
    \node[det, above=of d1,minimum size=25pt] (d2) {$\v{d}_2$} ; %
    \node[det, above=of d2,minimum size=25pt] (d3) {$\v{d}_3$} ; %
    \node[latent, above right=0.5cm and 0.4cm of d1] (z1) {$\v{z}_1$} ; %
    \node[latent, above right=0.5cm and 0.4cm of d2] (z2) {$\v{z}_2$} ; %
    \node[latent, above right=0.5cm and 0.4cm of d3] (z3) {$\v{z}_3$} ; %
    \node[det, right=0.5cm of z1,minimum size=25pt] (e1) {$\v{e}_{\theta,1}$} ; %
    \node[det, right=0.5cm of z2,minimum size=25pt] (e2) {$\v{e}_{\theta,2}$} ; %
	\edge[blue] {z3} {d3} ; %
	\edge {e2} {z2} ; %
	\edge[blue] {z2} {d2} ; %
	\edge {e1} {z1} ; %
	\edge {z1} {d1} ; %
	\edge {d3} {e2} ; %
	\edge {d2} {e1} ; %
	\edge[blue] {d3} {d2} ; %
	\edge {d2} {d1} ; %
	\edge {d1} {x} ; %
    \plate[inner sep=0.3cm] {plate1} {(x) (z1) (z2) (z3) (e1) (e2) (d1) (d2) (d3)} {\scalebox{1}{{$N$}}}; %
  }}
  \label{fig:gin_var}}}
 \begin{minipage}[c]{\columnwidth}
   \vspace{1mm}
\noindent\makebox[\columnwidth]{%
  \subfloat[Diagrammatic Representation\label{fig:diagram}]{
  \raisebox{-1em}{
  \begin{tikzpicture}[scale=0.4, every node/.style={scale=0.4},on grid,font=\fontsize{20}{20}\selectfont]]
    \begin{scope}[
    	yshift=10,every node/.append style={
    	    yslant=0.5,xslant=-1},yslant=0.5,xslant=-1
    	             ]
        \fill[white,fill opacity=.9] (-1,-1) rectangle (5,5);
        \draw[black,very thick] (-1,-1) rectangle (5,5);
        \draw[step=2.5mm, black] (-1,-1) grid  (5,5);
    	\draw[red, -, thick] (-1.2, -1) to (-1.2, 5);
        \draw[red](-2,2) node {${\mathsf{\sqrt{d_x}}}$};
    \end{scope}
    	
    \begin{scope}[
    	yshift=90,every node/.append style={
    	yslant=0.5,xslant=-1},yslant=0.5,xslant=-1
    	             ]
    	\fill[white,fill opacity=.9] (0,0) rectangle (5,5);
    	\draw[step=5mm, black] (0,0) grid (4,4);
    	\draw[black,very thick] (0,0) rectangle (4,4);
    	\draw[black,dashed] (0,0) rectangle (4,4);
    	\draw[red, -, thick] (-0.2,0) to (-0.2, 4);
        \draw[red](-0.8,2.8) node{{$\frac{\mathsf{\sqrt{d_x}}}{2}$}};

    \end{scope}
    \begin{scope}[
    	yshift=165,every node/.append style={
    	yslant=0.5,xslant=-1},yslant=0.5,xslant=-1
    	             ]
     	\fill[white,fill opacity=.95] (1,1) rectangle (3,3);
    	\draw[step=5mm, black] (1,1) grid (3,3);
    	\draw[black,very thick] (1,1) rectangle (3,3);
    	\draw[black,dashed] (1,1) rectangle (3,3);
        \draw[red, -, thick] (0.8,1) to (0.8, 3);
    	\draw[red](0.2,2.2) node{{$\frac{\mathsf{\sqrt{d_x}}}{4}$}};
    \end{scope}
    
    \begin{scope}[
    	yshift=230,every node/.append style={
    	yslant=0.5,xslant=-1},yslant=0.5,xslant=-1
    	             ]
    	\draw[step=10mm, black] (1.5,1.5) grid (2.5,2.5);
    	\draw[black,very thick] (1.5,1.5) rectangle (2.5,2.5);
    	\draw[black,dashed] (1.5,1.5) rectangle (2.5,2.5);
        \draw[red, -, thick] (1.3,1.5) to (1.3, 2.5);
    	\draw[red](0.7,1.8) node{{$\frac{\mathsf{\sqrt{d_x}}}{8}$}};
    \end{scope} %
    \draw[thick](6.8,2.35) node{$\mathbf{x}$};

    \draw[thick](4.8,5.2) node{$\mathbf{z}_1$};
    \draw[thick](2.8,7.8) node{$\mathbf{z}_2$};
    \draw[thick](1.8,10.1) node{$\mathbf{z}_3$};

\end{tikzpicture}}}}
\end{minipage}
 \caption{RRVQ--VAE with $L=3$, (a) variational posterior and (b) generative model, as defined in Eq~\eqref{eq:lvae_elbo}. Blue arrows indicate shared networks. For simplicity the codebooks are not represented. (c) is a diagrammatic representation of the model, showing the spatial arrangement of latents. We decrease the multiplicity by a factor of 4 at each layer.}
\vspace{-5mm}
\label{fig:modelgraphs}
\end{figure}

\subsection{Overall Model}
By combining Relaxed-Responsibility VQ with a hierarchical discrete VAE structure, we obtain our proposed model, a Relaxed-Responsibility Vector Quantised VAE (RRVQ--VAE).
See Fig~\ref{fig:modelgraphs} for a graphical representation of this model.
There is a deterministic chain in the inference network, the representations $\{\smash{\hat{\v{d}}}_\ell\}$.
Similarly, there is a deterministic downwards chain of representations ${\{\v{d}_\ell\}}$ in the generative model.
These representations enable the conditional structure given in Eqs~(\ref{eq:p_lad}-\ref{eq:q_lad}): that in the generative model we have an autoregressive structure over layers, and similarly that in the posterior each layer of latents is conditioned both on $\v{x}$ and on those above it in the hierarchy.
We choose to have a progressively smaller number of latent variables per layer as we ascend the hierarchy.
If we continue decreasing the number until the top-most latent is a single discrete variable, it is reasonable for us to place a uniform categorical prior over it.
Following continuous VAE models, including Ladder-VAEs~\citep{Sonderby2016}, ResNet-VAEs~\citep{Kingma} and BIVA~\citep{Maaloe2019}, we enforce weight sharing between the generative and inference networks, indicated by \textcolor{Blue}{blue} arrows.

\section{Experiments}
We train our model for image reconstruction on CIFAR-10, SVHN and CelebA.
We train very deep models with $L=32$ layers, as well as smaller $L=5$ models for visualisation and ablation studies.
For each, the models for CIFAR-10 and SVHN have identical specification, with some small changes for CelebA due to the different image size.
We implement these models using fully convolutional networks composed of ResNet blocks.

The number of latent variables per layer decreases as we ascend the hierarchy, as represented in Fig \ref{fig:modelgraphs}(c).
For the $L=32$ model we decrease the number of latents by a factor of 4 every 8 layers, forming 4 blocks each of decreasing numbers of latents.
For $L=5$ models we reduce the number of latents by a factor of 4 each layer.
Each layer of latent variables has its own pair of codebooks for means and diagonal covariances.
For further model description and implementation details, see Appendix~\ref{app:implementation}.

\subsection{Numerical Results}
\begin{table}[h]
  \caption{Bits Per Dim Results: Comparison of our model, RRVQ--VAE, to rVQ--VAEs in bits-per-dim (bpd) for train \& test sets -- lower better. We also benchmark against VIMCO-trained discrete VAEs~\citep{VQVAE} and FouST-trained models with binary latents and $L=1$ or $L=4$ layers~\citep{Pervez2020}. For additional context we also give values for hierarchical Spatial-VAEs, the Gaussian-latent-variable version of our models.}
  \label{tab:results}
  \centering
  \begin{sc}
  \begin{tabular}{lcc}
    \toprule
    Model              & Test bpd & Train bpd\\
    \midrule
    \multicolumn{3}{c}{CIFAR-10}\\
    \cmidrule(r){1-3}
    VIMCO & 5.14 & - \\
    rVQ--VAE   & 4.77 & 4.87\\
    FouST, $L=4$ & 4.16 & - \\
    FouST, $L=1$ & 4.02 & - \\
    RRVQ--VAE, $L=32$   & \textbf{3.94}&\textbf{3.81} \\
     \cmidrule(r){1-3}
Spatial--VAE, $L=32$   & 3.55 & 3.49 \\
    \midrule
    \multicolumn{3}{c}{SVHN}\\
    \cmidrule(r){1-3}
    rVQ--VAE & 3.73 & 4.17   \\
    RRVQ--VAE, $L=32$ & \textbf{2.30}&\textbf{2.52}\\
     \cmidrule(r){1-3}
Spatial--VAE, $L=32$   & 1.94 & 2.07 \\
    \midrule
    \multicolumn{3}{c}{CelebA}\\
    \cmidrule(r){1-3}
    rVQ--VAE & 5.31 & 5.31\\
    RRVQ--VAE, $L=32$ & \textbf{2.97}&\textbf{2.97}\\
     \cmidrule(r){1-3}
Spatial--VAE, $L=32$   & 2.54 & 2.58 \\
    \bottomrule
  \end{tabular}
  \end{sc}
\end{table}
\begin{figure*}[t!]
    \centering
    \foreach \i [evaluate=\i as \I using int(\i+1)] in {0,...,4}{%
        \subfloat[$\ell = \I$]{\includegraphics[width=0.19\textwidth]{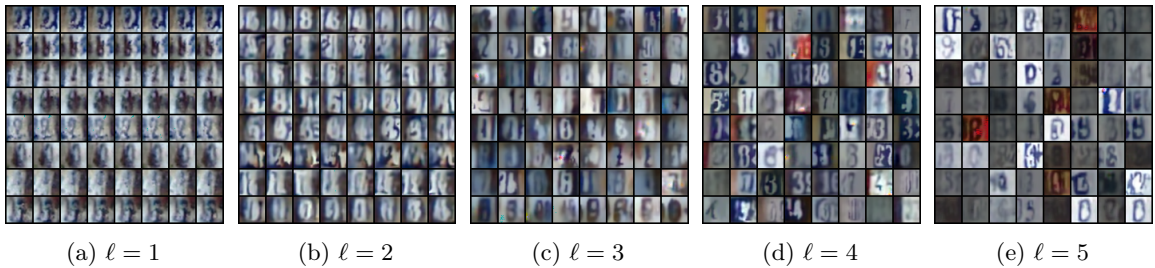}}
        \hspace*{1mm}
    }%
    \hspace*{-2mm}
    \caption{Layerwise sampling in 5 layer RRVQ--VAE trained on SVHN. Note that layer $\ell=2$ seems to represent digit identity: resampling in this layer changes digit identity while keeping the rest of the image roughly the same.}
    \label{fig:interpret}
\end{figure*}
We show in Table~\ref{tab:results} numerical results from our $L=32$ models, benchmarked against rVQ--VAEs and various baselines.
We measure the bits-per-dim (bpd) for the training and test set (using non-relaxed categorical distributions).

From one end of the spectrum, our baseline is a non-hierarchical rVQ--VAE trained with a set of uniform categorical priors.
From the other end, we also train the continuous version of our hierarchical model, where we have Gaussian latent variables rather than codebook embeddings at each latent position.
This is the hierarchical version of the Spatial-VAE of \S\ref{sec:sampling}.
This is very similar to a ResNet VAE, but without inverse autoregressive flows \citep{Kingma}.
These results show clearly the benefits our approach brings to VQ--VAEs, from the improved values reached of test and train bits-per-dim.
Our models help close the gap between discrete latent variable models and those with continuous latent variables.

Further to this, Fig~\ref{fig:recons} shows reconstructions and Fig~\ref{fig:samples} ancestral samples for our models and baselines.
Our RRVQ--VAE model achieves reasonable sampling quality without having to train a second post-hoc model over the learnt embeddings.
Though our model's sampling quality is not as good as VQ–VAE models with large, post-trained, autoregressive priors, we provide a more efficient sampling method that only requires a single pass.
A hierarchical model with autoregressive priors requires as many internal forward passes as it has latent variables, creating very expensive sampling procedures. 
Our results are the first for a  hierarchical probabilistic deep generative model with discrete latent variables trained in a unified manner.
\subsection{Analysis of Samples and Representations}

In addition to the 32 layer models, we also trained a 5 layer version for simpler plotting and analysis.
Fig~\ref{fig:interpret} demonstrates the effect each hierarchical layer has on the final draws when we train this smaller model on SVHN.
For this figure we sample repeatedly (plotted along each row) in each layer, conditioned on sampled value of all the layers above (each set of these values shown in a different row).
We then propagate deterministically down through the layers below -- we take the mode for each subsequent layer.

The resulting latent structure indicates that different layers are representing different aspects of the data: the layers showing a degree of separation in their purpose.
For instance it might seem that $\ell =2$ describes digit identity.
To verify this, we trained simple convnets on the embeddings from the $L=1$ rVQ baseline and those from the $\ell=2$ layer of this model.
From the results in Table~\ref{tab:identity_results} we can see that digit identity is concentrated in the $\ell=2$ layer.
The rVQ--VAE provides good reconstructions, so its embeddings do, necessarily, encode digit identity, yet the $\ell=2$ layer (which does not encode all information for reconstruction) makes digit identity easier for a convnet to ascertain.
We also experiment with using these $L=5$ models for compression -- see Appendix~\ref{app:compress}.
\begin{table}[h]
  \caption{Digit Classification Results: SVHN embeddings from the $\ell=2$ layer of our $L=5$ model and from an $L=1$ baseline were each used to train small convnets. We give the test-set accuracy over 4 runs.}
  \label{tab:identity_results}
  \centering
  \begin{sc}
  \begin{tabular}{lc}
    \toprule
    Model              & Test Set Acc\\
    \midrule
    rVQ--VAE   &$0.578\pm0.008$\\
    RRVQ--VAE, $L=5$    &$\mathbf{0.626\pm0.007}$ \\
    \bottomrule
  \end{tabular}
  \end{sc}
\end{table}

\subsection{Ablation Study}
\label{sec:ablate}
How does our approach compare to other possible hierarchical extensions of rVQ--VAEs?
For the $L=5$ models we trained various ablations of our proposal: with or without a learnt codebook of covariances; and with the generative model represented either via a Relaxed-VQ lookup or directly outputting a (log) probability over embeddings.
Thus all of these have an ELBO as in Eq~\eqref{eq:lvae_elbo}, but vary in how we parameterise $p$ and $q$.

In Table~\ref{tab:ablate} we show the test and train bpd obtained for these hierarchical discrete VAEs.
\begin{table}[h]
  \caption{Ablation Study for $L=5$ models on CIFAR-10 and SVHN: We show the train and test bits-per-dim We can have: the generative model log probabilities directly output by a net ($p$: \textit{Direct-Cat}) or parameterised using responsibilties in the embedding space ($p$: \textit{Embed-Cat}), and we can learn a codebook of diagonal covariances for the responsibilities (\textit{$\sigma$ learnt}) or have them all fixed to one (\textit{$\sigma=1$}). RRVQ is when we have \textit{Embed-Cat in $p$} and \textit{$\sigma$ learnt}. Note that Direct-Cat with $\sigma$ learnt is unstable during training for SVHN.}
  \label{tab:ablate}
    \centering
    \begin{sc}
  \begin{tabular}{rrlrl}
      \toprule
 $p$:   & \multicolumn{2}{c}{Direct-Cat}   & \multicolumn{2}{c}{Embedding-Cat} \\
    \midrule
    \multicolumn{5}{c}{CIFAR-10}\\
    \cmidrule(r){1-5}
    \multirow{2}{*}{$\sigma = 1$} & Train: & 5.00 & Train: & 5.06 \\
    & Test:& 5.05 & Test:& 5.11 \\
        \cmidrule(r){2-5}
    \multirow{2}{*}{$\sigma$ learnt} & Train:& 5.08 & Train:& \textbf{4.40} \small{\textit{(RRVQ)}} \\
    & Test:& 5.10 & Test:& \textbf{4.65} \small{\textit{(RRVQ)}} \\
    \midrule
    \multicolumn{5}{c}{SVHN}\\
    \cmidrule(r){1-5}
    \multirow{2}{*}{$\sigma = 1$} & Train:& 3.44 & Train:& 3.51 \\
    & Test:& 3.32 & Test:& 3.41 \\
        \cmidrule(r){2-5}
    \multirow{2}{*}{$\sigma$ learnt} & Train:& -- & Train:& \textbf{3.02} \small{\textit{(RRVQ)}}\\
    & Test:& -- & Test:& \textbf{2.96} \small{\textit{(RRVQ)}} \\
      \bottomrule
  \end{tabular}
  \end{sc}
\end{table}
We can see that \textit{$\sigma=1$} with \textit{direct probabilities in $p$} (the top-left corner results for each dataset), arguably the most naive approach, is outperformed by $\approx$ half of a bpd by full RRVQ (bottom-right).
Interesting, either of the two changes made to obtain RRVQ made in isolation lead either no substantive change in performance (if anything, slight degradation) or rendered training so unstable that it was impossible to obtain a result.
Clearly there is a synergistic property here, that the these two changes together lead to improved performance of the models.

\section{Related Work}
VQ--VAEs have been extended to the two- and three-layer case~\citep{Razavi2019}, with large, powerful autoregressive models subsequently trained as priors to then sample from, producing draws competitive with the state of the art when combined with a classifier-based accept-reject algorithm.
Various recent papers have worked towards hierarchical discrete VAEs that eschew the training of priors as auxiliary models.

One recent work trains layers of discrete latent variables in various hierarchical arrangements on MNIST and Fashion-MNIST~\citep{Lievin2019}, building on variational memory addressing methods~\citep{Bornschein2017}.
In Hierarchical Quantised Autoencoders \citep{Williams2020}, much like in the original VQ--VAE paper, a sequential training pipeline is proposed.
Here rVQ--VAEs are trained one at a time, with the first trained on the dataset and each subsequent sub-model trained on sampled values of the latents from the one below.
This gives a Markovian structure, both in the generative and inference networks.
\citet{Pervez2020} performs inference over binary latents using a novel Gumbel-Softmax-derived method, but there hierarchy in the model seems to harm rather than help performance.

Methods have been developed to perform bits-back coding~\citep{Frey1996} using the learnt representations of VAEs~\citep{Townsend2019}, including for hierarchical VAEs~\citep{Townsend2019a}.
In these methods the latents are continuous during training, with the space then subsequently bucketed.
Recently flow-based models~\citep{Dinh2015, Dinh2017, deepflows} have been extended to handle discrete variables~\citep{Hoogeboom2019, Tran2019}.

Recently VAEs with conditionally-Gaussian latents have enjoyed a resurgence, with VAEs with the same probabilistic `wiring' as studied here, that of \citet{Kingma}, with deep hierarchies of latents obtaining state-of-the-art performance~\citep{Vahdat2020,Child2020}.

As discussed in \S\ref{sec:rrvq}, vector quantisation has close links to mixture models and mixtures of experts~\citep{Jacobs1991}.
Historically it has been known that stochastic relaxations of vector quantisation offer various benefits compared to deterministic assignment, and that they are equivalent to certain classes of mixture models~\citep{Hinton1994}.

Vector quantisation can be thought of as inference on a Voronoi partition \nbcite[\S5]{Sack2000}.
Our distributions are the responsibilities from a mixture model with learnt variances, so deterministic RRVQ would result in Mahalanobis-distance Voronoi partitions.
\section{Conclusion}
We have presented a novel parameterisation for stochastic Vector Quantisation, Relaxed-Responsibility Vector Quantisation. RRVQ learns a codebook of variances alongside the codebook of means, using the responsibilities under the Gaussian mixture model represented by those quantities to define discrete distributions, both within the approximate posterior using for inference and in the forward model.

We then use this is as a building block to develop a novel variety of hierarchical discrete VAE, Relaxed-Responsibility Vector-Quantised VAEs.
RRVQ--VAEs are the first unified probabilistic deep generative models with hierarchies of discrete latent variables to be trained end-to-end on the datasets studied.

RRVQ--VAEs are highly expressive; their hierarchy of representations separate out different aspects of the data.
The capacity and flexibility of the models is demonstrated by the fact they produce samples without training a secondary autoregressive generative models over posterior latent samples.
Further, they avoid the large number of forward passes that a hierarchical model of that form would need to produce a single sample.
We hope that this work inspires further research into discrete hierarchical variational autoencoders, with the aim of completely closing the gap between hierarchical discrete VAEs and those with continuous latent variables.

\newpage

\newpage
\clearpage
\bibliographystyle{apalike2}
\bibliography{references.bib}

\newpage
\appendix
\setcounter{equation}{0}
\setcounter{figure}{0}
\renewcommand\theequation{\thesection.\arabic{equation}}
\renewcommand\thefigure{\thesection.\arabic{figure}}
\setcounter{table}{0}
\renewcommand{\thetable}{\thesection.\arabic{table}}
 \onecolumn
\clearpage
  \hsize\textwidth
  \linewidth\hsize {\centering
  {\Large\bfseries Appendix for Relaxed-Responsibility Hierarchical Discrete VAEs \par}}
 
\section{Relaxed Responsibility Vector Quantisation from a Mixture Model}
\label{app:A}
To gain more insight into VQ-derived models, we can take a hierarchical discrete VAE and within it promote the embedding outputs $\vv{e}=\{\v{e}_1,..,\v{e}_L\}$ to probabilistic variables.
In doing this we obtain a hierarchical Gaussian mixture model, where each layer is itself a set of Gaussian mixture latent variables:
\begin{equation}
p_\theta(\v{x},\vv{z},\vv{e}) = p_\theta(\v{x}|\vv{e})p_\theta(\vv{e},\vv{z}) = p_\theta(\v{x}|\vv{e})\prod_{\ell=1}^{L-1} [p(\v{e}_\ell|\v{z}_{\ell})p_\theta(\v{z}_\ell|\v{e}_{>\ell})]p(\v{z}_L)
\end{equation}
where 
\begin{equation}
p(\v{e}_\ell|\v{z}_{\ell}) = \prod_{m=1}^M\mathcal{N}(\v{e}^m_\ell|\v{\mu}=\v{E}_{\mu,\ell}\v{z}^m_{\ell}, \v{\Sigma}=\v{E}_{\Sigma,\ell}\v{z}^m_\ell)\end{equation} and \begin{equation}p_\theta(\v{z}_\ell|\v{e}_{>\ell})=\prod_{m=1}^M\cat(\v{z}_{m}|\pi_\theta^{m}(\v{e}_{>\ell})).\end{equation}
The posterior is given by \begin{equation}q_\phi(\vv{z},\vv{e}|\v{x})=q_\phi(\v{z}_L|\v{x})q_\phi(\v{e}_L|\v{z}_L)\prod_{\ell=1}^{L-1}q_\phi(\v{z}_\ell|\v{e}_{>\ell},\v{x})q_\phi(\v{e}_\ell|\v{z}_\ell).\end{equation}

We can obtain our model as a restricted version of this.
Our intent is to bottleneck our representations through a set of discrete latent variables.
Thus we choose $q_\phi(\v{e}_\ell|\v{z}_\ell) =p_\theta(\v{e}_\ell|\v{z}_\ell) = \delta(\v{e}_\ell - \v{E}_{\mu,\ell}\v{z}_\ell)$, where $\delta(\cdot)$ is the Dirac delta function.
This gives us an ELBO of the form
\begin{align}
    \ELBO(\v{x})=&  \expect_{\vv{z}\sim q} \log p_\theta(\v{x}|\vv{z}) -\sum_{\ell=1}^{L-1} \expect_{\v{z}_{>\ell}\sim q}\KL(q_\phi(\v{z}_\ell|\v{z}_{>\ell}, \v{x})||p_\theta(\v{z}_\ell|\v{z}_{>\ell})) - \KL(q_\phi(\v{z}_L|\v{x})||p(\v{z}_L)),
\end{align}
where we have changed the likelihood to depend on $\vv{z}$, as the $\vv{e}$ it depended on is now deterministic given $\vv{z}$.
This mirrors our original notation, where one writes $p_\theta(\v{x}|\v{z})$ and $\v{z}$ implicitly looks-up the codebook embeddings inside the likelihood.
If we then we choose Eqs~(\ref{eq:rr_q},~\ref{eq:rr_p}) to parameterise the inference and generative models of each $\v{z}$, we thus obtain our RRVQ--VAE.

\section{Interpreting Discrete Hierarchical VAEs as Learning a Series of Reconstructions}
\label{app:unity}
We can expand each $\KL$ in Eq~\eqref{eq:lvae_elbo} as a cross entropy and an entropy: $\KL(q||p) = \ent(q||p) - \ent(q)$.
The ELBO for this model can then be written as
\begin{align}
    \ELBO(\v{x}) =& \expect_{\vv{z}\sim q}[ \ent(q(\v{x})||p_\theta(\v{x}|\vv{z}))] \nonumber \\ & \phantom{{}=1}-\sum_{\ell=1}^{L-1} \expect_{\v{z}_{>\ell}\sim q}[\ent(q_\phi(\v{z}_\ell|\v{z}_{>\ell}, \v{x})||p_\theta(\v{z}_\ell|\v{z}_{>\ell}))- \ent(q_\phi(\v{z}_\ell|\v{z}_{>\ell}, \v{x}))]
     \nonumber \\
     &\phantom{{}=1}- \ent(q_\phi(\v{z}_L|\v{x})||p(\v{z}_L)) + \ent(q_\phi(\v{z}_L|\v{x})),
    \label{eq:lvae_elbo_xent}
\end{align}

where $q(\v{x})$ is the per-datapoint empirical distribution (we view a datapoint as a set sub-pixels) of one-hot discrete distributions.

If the likelihood $p_\theta(\v{x}|\vv{z})$ is itself a set of discrete distributions, then in a hierarchical discrete VAE the latent layers and the likelihood term all provide to the ELBO cross-entropy terms between discrete distributions, with then the entropy of each latent posterior acting as regularisers.
If that is that case, then during training we are, in effect, requiring our model to build a series of representations $\vv{z}$, all of which are scored under local objectives of the same form as how we score the reconstruction of our datapoint under our likelihood.

\begin{figure}[h]
    \centering
    \includegraphics[width=0.3\textwidth]{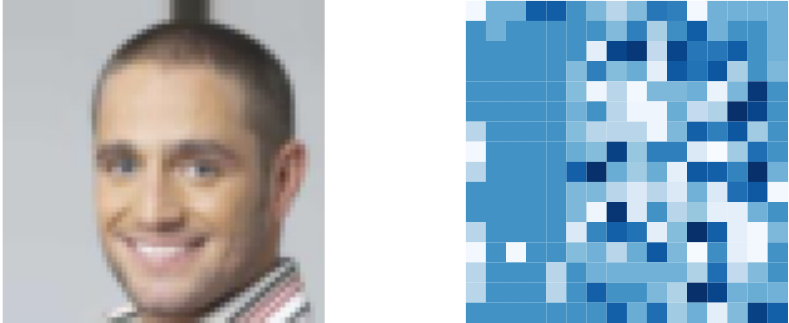}\qquad\qquad\qquad
    \includegraphics[width=0.3\textwidth]{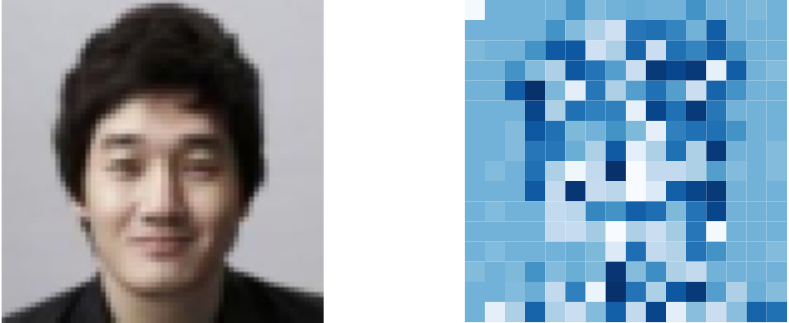}
    \caption{For two input images from CelebA we plot them and their $\v{z}_1$ representations from a RRVQ--VAE, colouring the indexes using the norm of the corresponding codebook mean.}
    \label{fig:recon_code}
\end{figure}

One might expect that the embedding of images when plotted as an image looks somewhat like a compressed version of the input data, up to the arbitrary indexing of the discrete latents.
This is a weak effect, but can be seen somewhat in Fig \ref{fig:recon_code} below for two datapoints from CelebA.
For each, the background is being encoded mostly using a single codebook index, which means that the person can be seen segmented out spatially in the latent representation in the first layer.

\section{Details of Model Architecture}
\label{app:implementation}
\begin{figure}[h!]
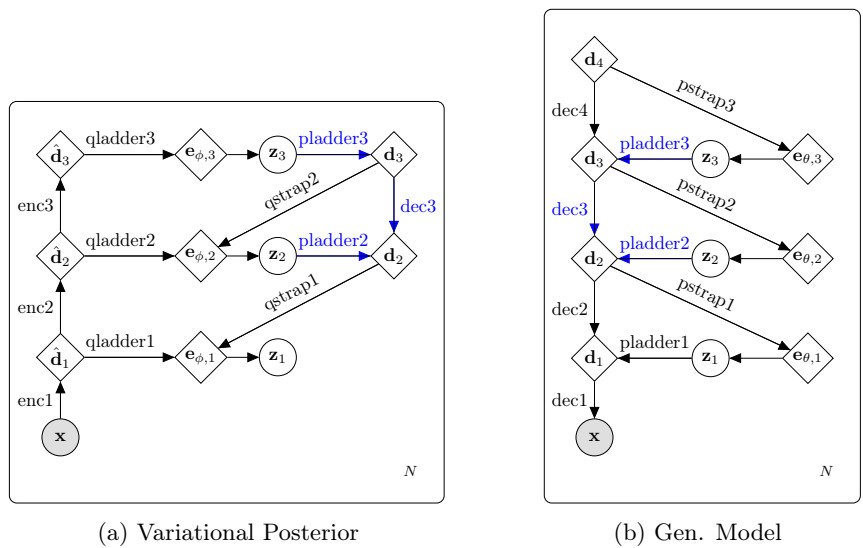

\noindent\makebox[\textwidth]{%
\subfloat[Variational Posterior]{\scalebox{0.7}{\raisebox{0ex}{
\tikz{ %
      \node[obs] (x) {$\v{x}$} ; %
    \node[det, above=0.7cm of x,minimum size=25pt] (d1) {$\hat{\v{d}}_1$} ; %
    \node[det, above=of d1,minimum size=25pt] (d2) {$\hat{\v{d}}_2$} ; %
    \node[det, above=of d2,minimum size=25pt] (d3) {$\hat{\v{d}}_3$} ; %
    \node[det, right=1.7cm of d1] (e1) {$\v{e}_{\phi,1}$} ; %
    \node[det, right=1.7cm of d2] (e2) {$\v{e}_{\phi,2}$} ; %
    \node[det, right=1.7cm of d3] (e3) {$\v{e}_{\phi,3}$} ; %
    \node[latent, right=0.6cm of e1] (z1) {$\v{z}_1$} ; %
    \node[latent, right=0.6cm of e2] (z2) {$\v{z}_2$} ; %
    \node[latent, right=0.6cm of e3] (z3) {$\v{z}_3$} ; %
    \node[det, right=1.4cm of z3,minimum size=25pt] (d_3) {$\v{d}_3$} ; %
    \node[det, right=1.4cm of z2,minimum size=25pt] (d_2) {$\v{d}_2$} ; %
	\edge {x} {d1} ; %
	\edge {d1} {d2, e1} ; %
	\edge {d2} {d3, e2} ; %
	\edge {d3} {e3} ; %
	\edge {e1} {z1} ; %
	\edge {e2} {z2} ; %
	\edge {e3} {z3} ; %
	\edge[blue] {z3} {d_3} ; %
	\edge {d_3} {e2} ; %
	\edge {d_2} {e1} ; %
	\edge[blue] {d_3} {d_2} ; %
	\edge[blue] {z2} {d_2} ; %
    \draw (x)  --  (d1) 
        node [midway,left](enc1){enc1};
    \draw (d1)  --  (d2) 
        node [midway,left](enc2){enc2};
    \draw (d2)  --  (d3) 
        node [midway,left](enc3){enc3};
    \draw (d1)  --  (e1) 
        node [pos=0.4,above](qlad1){qladder1};
    \draw (d2)  --  (e2) 
        node [pos=0.4,above](qlad2){qladder2};
    \draw (d3)  --  (e3) 
        node [pos=0.4,above](qlad3){qladder3};
    \draw (z3)  --  (d_3) 
        node [blue, pos=0.5,above,sloped](plad3){pladder3};
    \draw (z2)  --  (d_2) 
        node [blue, pos=0.5,above,sloped](plad2){pladder2};
    \draw (d_3)  --  (e2) 
        node [midway, above, sloped](qstrap2){qstrap2};
    \draw (d_2)  --  (e1) 
        node [midway, above, sloped](qstrap1){qstrap1};
    \draw (d_3)  --  (d_2) 
        node [blue, midway, right](dec3){dec3};
    \plate[inner sep=0.5cm] {plate1} {(x) (z1) (z2) (z3) (e1) (e2) (e3) (d1) (d2) (d3) (d_2) (d_3)} {\scalebox{1}{{$N$}}}; %
  }}
  \label{fig:gout_aoo}
  }}
\hspace*{9mm}
\subfloat[Gen. Model]{\scalebox{0.7}{
\tikz{
    \node[obs] (x) {$\v{x}$} ; %
    \node[det, above=0.7cm of x,minimum size=25pt] (d1) {$\v{d}_1$} ; %
    \node[det, above=of d1,minimum size=25pt] (d2) {$\v{d}_2$} ; %
    \node[det, above=of d2,minimum size=25pt] (d3) {$\v{d}_3$} ; %
    \node[det, above=of d3,minimum size=25pt] (d4) {$\v{d}_4$} ; %
    \node[latent, right=1.4cm of d1] (z1) {$\v{z}_1$} ; %
    \node[latent, right=1.4cm of d2] (z2) {$\v{z}_2$} ; %
    \node[latent, right=1.4cm of d3] (z3) {$\v{z}_3$} ; %
    \node[det, right=1cm of z1,minimum size=25pt] (e1) {$\v{e}_{\theta,1}$} ; %
    \node[det, right=1cm of z2,minimum size=25pt] (e2) {$\v{e}_{\theta,2}$} ; %
    \node[det, right=1cm of z3,minimum size=25pt] (e3) {$\v{e}_{\theta,3}$} ; %
	\edge[blue] {z3} {d3} ; %
	\edge[blue] {z2} {d2}; %
	\edge {z1} {d1} ; %
	\edge {e3} {z3} ; %
	\edge {e2} {z2} ; %
	\edge {e1} {z1} ; %
	\edge {d4} {e3} ; %
	\edge {d3} {e2} ; %
	\edge {d2} {e1} ; %
	\edge {d4} {d3} ; %
	\edge[blue] {d3} {d2} ; %
	\edge {d2} {d1} ; %
	\edge {d1} {x};
    \draw (d1)  --  (x) 
        node [midway,left](dec1){dec1};
    \draw (d2)  --  (d1) 
        node [midway,left](dec2){dec2};
    \draw (d3)  --  (d2) 
        node [blue, midway,left](dec3){dec3};
    \draw (d4)  --  (d3) 
        node [midway,left](dec4){dec4};
    \draw (z3)  --  (d3) 
        node [blue, pos=0.5,above,sloped](plad3){pladder3};
    \draw (z2)  --  (d2) 
        node [blue, pos=0.5,above,sloped](plad2){pladder2};
    \draw (z1)  --  (d1) 
        node [pos=0.5,above,sloped](plad1){pladder1};
    \draw (d4)  --  (e3) 
        node [midway, above, sloped](pstrap3){pstrap3};
    \draw (d3)  --  (e2) 
        node [midway, above, sloped](pstrap2){pstrap2};
    \draw (d2)  --  (e1) 
        node [midway, above, sloped](pstrap1){pstrap1};
    \plate[inner sep=0.5cm] {plate1} {(x) (z1) (z2) (z3) (e1) (e2) (d1) (d2) (d4)}
    {\scalebox{1}{{$N$}}};
    }}
    }}
 \caption{RRVQ--VAE with $L=3$ as an example. (a) the variational posterior and (b) generative model, as defined in Eq~\eqref{eq:lvae_elbo}. Blue arrows indicate shared networks. For simplicity the codebooks are not represented. Each labelled arrow corresponds to a network, described below.
}
 \label{fig:modelgraphs_app}
\end{figure}

The basic structure our network implementation is that of a ResNet VAE \citep{Kingma}.
Now we describe the structure of each variety of network inside our model.
enc1/dec1 are convolutions/transposed convolutions that down/upscale their inputs using a stride of 2.
All the other subnetworks of the enc/dec deterministic backbones are each implemented as a single resnet block -- each dec\_ using a transposed convolution internally.
When the mappings between two layers of latent variables requires a resizing, the identity path of the network performs a differentiable rescaling operation.

The networks qladder\_ map from the backbone of encoders to the embedding space, and pladder\_ map from the embedding space to the backbone of decoders.
Each of these are each implemented as a single convolutional layer.
The networks qstrap\_ and pstrap\_ too are each implemented as a single convolutional layer, and carry out upscaling using a stride of 2.
They output in the embedding space.
The embeddings used to define each layer's posterior distribution are the sum of the outputs of that layer's qstrap\_ and qladder\_ networks, and the embeddings used for the generative model's internal conditional probabilities are simply the outputs of each pstrap\_.

For the $L=32$ runs the backbones have 256 channels, and the $\v e$ representations are 32 dimensional.
Each layer's codebooks each hold 256 embeddings.
The likelihood function is the same discretised logistic likelihood as in \cite{Kingma}.
As in \cite{Kingma}, we use weight normalisation, ELU activations and free-bits regularisation.

The top-most latent variable in the generative model can be set to be uniform over embeddings, or can be parameterised by a similar procedure as for the rest via a $\v{d}_L$ that is a learnable parameter (rather than itself the output of a network).
See Fig~\ref{fig:modelgraphs_app} for a representation of this -- here for $L=3$ we have $\v{d}_4$ in the generative model parameterising $p_\theta(\v{z}_3)$.

When training with just $L=5$ layers, as opposed to 32, it is as if we remove the corresponding intermediate latent variables along with their ladder\_, strap\_ networks, so now the enc\_ and dec\_ networks are composed of 4 resnet blocks between latents.
We also promote the remaining ladder\_, strap\_ networks to themselves be composed of 4 resnet blocks.

We train using AdaMax with batch size 64 and an initial learning rate that we decay on plateau, multiplying by 0.8 when there has been no decrease in the test set ELBO for 20 (SVHN + CIFAR-10)/5 (CelebA) epochs, down to a minimum of $5\times 10^{-5}$.
The initial learning rate is $2\times 10^{-3}$.
We train with for up to 500 (SVHN + CIFAR-10)/160 (CelebA) epochs or until convergence.
We used Azure VMs with NVIDIA M60 GPUs to train our models -- using a single M60 to train a model takes $\approx$ 1 week for SVHN and CIFAR10.
For the CelebA multi-GPU training is necessary.

\newpage
\section{Worst-Case Entropy of rVQ and Softmax-parameterised Discrete Distributions}
\label{app:entropy}
\subsection{Proof of Theorem 1}
\label{app:theorem1}
\begin{proof}
Our distribution of interest is a rVQ distribution, ie Eq~\eqref{eq:gssoft_2}, where we have the worst possible arrangement of our $K$-member codebooks -- the arrangement that leads to the minimum possible entropy, and we also assume the worst possible positions of the embedding vector $\v e$.
The arrangement that leads to this is having all but one of the codebook vectors at one point and a single codebook separated a distance $\delta$ from them, with the embedding vector $\v e$ lying along the line defined by those two positions a distance $d$ from the outlier codebook vector and $d+\delta$ from the remaining $K-1$ codebook vectors.
We note that this arrangement is closely related to that considered in \citet{Beyer1998}, \S~3.5.2.

This gives us a distribution $p(\v z|\v{\pi})$, where
\begin{align}
   \pi^i & = \left\{ \,
\begin{IEEEeqnarraybox}[][c]{l?s}
\IEEEstrut
\frac{1}{Z}\exp{\left(-\frac{1}{2}d^2\right)} & if $i$ = 1 \\
\frac{1}{Z}\exp{\left(-\frac{1}{2}(d+\delta)^2\right)} & otherwise.
\IEEEstrut
\end{IEEEeqnarraybox}
\right.
\end{align}
and
\begin{equation}
    Z = \exp{\left(-\frac{1}{2}d^2\right)} + (K-1)\exp{\left(-\frac{1}{2}(d+\delta)^2\right)}
\end{equation}

The entropy of this discrete distribution is thus:
\begin{align}
    \mathcal{H}_\mathrm{rVQ} &= -\sum_{i=1}^K \pi^i \log \pi^i \\
    &= -\frac{\exp{\left(-\frac{1}{2}d^2\right)}}{Z}\log\left(\frac{\exp{\left(-\frac{1}{2}d^2\right)}}{Z}\right) - (K-1)\frac{\exp{\left(-\frac{1}{2}(d+\delta)^2\right)}}{Z}\log\left(\frac{\exp{\left(-\frac{1}{2}(d+\delta)^2\right)}}{Z}\right)\\
    &= -\frac{\exp{\left(-\frac{1}{2}d^2\right)}}{Z}\left(-\frac{1}{2}d^2 - \log Z +(K-1)\exp{\left(-\frac{1}{2}\delta^2-\delta d\right)}\left(-\frac{1}{2}\left(d+\delta\right)^2-\log Z\right)\right)\label{eq:h_form_vq}
\end{align}
Now let us consider the value of this in the limit of large $d$, $d\gg\delta$.
First, let us expand $\frac{\exp{\left(-\frac{1}{2}d^2\right)}}{Z}$ using the first order expansion $(1+x)^{-1}\approx 1-x$ for $|x|\ll 1$.
\begin{align}
    \frac{\exp{\left(-\frac{1}{2}d^2\right)}}{Z} &= \frac{\exp{\left(-\frac{1}{2}d^2\right)}}{\exp{\left(-\frac{1}{2}d^2\right)} + (K-1)\exp{\left(-\frac{1}{2}(d+\delta)^2\right)}} \\
    &= \frac{1}{1 + (K-1)\exp{\left(-\frac{1}{2}(\delta^2 + 2\delta d)\right)}}\\
    &= 1 - (K-1)\exp{\left(-\frac{1}{2}(\delta^2 + 2\delta d)\right)} +O\left(\exp\left(-\frac{1}{2}(\delta^2 + 2\delta d)\right)^2\right). \label{eq:z_vq_expand}
\end{align}
Second, let us expand $\log Z$ using the first order expansion $\log(1+x)\approx x$ for $|x|\ll 1$.
\begin{align}
    \log Z &=\log\left(\exp{\left(-\frac{1}{2}d^2\right)} + (K-1)\exp{\left(-\frac{1}{2}(d+\delta)^2\right)}\right)\\
    &=\log\left(\exp{\left(-\frac{1}{2}d^2\right)}\left( 1+(K-1)\exp{\left(-\frac{1}{2}(\delta^2 + 2\delta d)\right)}\right)\right)\\
    &=-\frac{1}{2}d^2 + \log\left(1+(K-1)\exp{\left(-\frac{1}{2}(\delta^2 + 2\delta d)\right)}\right)\\
    &=-\frac{1}{2}d^2 + (K-1)\exp{\left(-\frac{1}{2}(\delta^2 + 2\delta d)\right)} +O\left(\exp\left(-\frac{1}{2}(\delta^2 + 2\delta d)\right)^2\right). \label{eq:logz_vq_expand}
    \end{align}
Taking Eqs~(\ref{eq:z_vq_expand},\ref{eq:logz_vq_expand}) and subbing back into Eq~\eqref{eq:h_form_vq}, we get
\begin{align}
    \mathcal{H}_\mathrm{rVQ} &=\left[1 - (K-1)\exp{\left(-\frac{1}{2}(\delta^2 + 2\delta d)\right)}\right]\bigg[ (K-1)\exp{\left(-\frac{1}{2}(\delta^2 + 2\delta d)\right)}\left(1+\frac{1}{2}(\delta+d)^2-\frac{1}{2}d^2\right)\nonumber\\
    &\hspace{18em}+O\left(\exp\left(-\frac{1}{2}(\delta^2 + 2\delta d)\right)^2\right)\bigg]\\
    &=(K-1)\exp{\left(-\frac{1}{2}(\delta^2 + 2\delta d)\right)}\left(1+\frac{1}{2}(\delta^2 + 2\delta d)\right)+O\left(\exp\left(-\frac{1}{2}(\delta^2 + 2\delta d)\right)^2\right).
\end{align}
Giving us, to first order in $\exp\left(-\frac{1}{2}(\delta^2 + 2\delta d)\right)$,
\begin{equation}
    \mathcal{H}_\mathrm{rVQ} \approx (K-1)\exp{\left(-\frac{1}{2}(\delta^2 + 2\delta d)\right)}\left(1+\frac{1}{2}\left(\delta^2 + 2\delta d\right)\right)
    \label{eq:rvq_approx_h}
\end{equation}
as required.
\end{proof}

\subsection{Proof of Theorem 2}
\label{app:theorem2}

\begin{proof}
Our distribution of interest is a discrete distribution defined as a softmax of $K$ raw logits, where we have the worst possible arrangement of the logit outputs -- the arrangement that leads to the minimum possible entropy.
The arrangement that leads to this is having all but one of the logits take one value $c$ and a single logit taking the value $c+\ell$, $\ell>0$.

This gives us a distribution $p(\v z|\v{\pi})$, where
\begin{align}
   \pi^i & = \left\{ \,
\begin{IEEEeqnarraybox}[][c]{l?s}
\IEEEstrut
\frac{1}{Z}\exp{\left(c+\ell\right)} & if $i$ = 1 \\
\frac{1}{Z}\exp{\left(c\right)} & otherwise.
\IEEEstrut
\end{IEEEeqnarraybox}
\right.
\end{align}
and
\begin{equation}
    Z = \exp{\left(c+\ell\right)} + (K-1)\exp{\left(c\right)}
\end{equation}

The entropy of this discrete distribution is thus:
\begin{align}
    \mathcal{H}_\mathrm{softmax} &= -\sum_{i=1}^K \pi^i \log \pi^i \\
    &= -\frac{\exp{\left(\ell+c\right)}}{Z}\left(\ell + c -\log Z\right) - (K-1)\frac{\exp{\left(c\right)}}{Z}\left(c-\log Z\right)\label{eq:h_form_softmax}
\end{align}
Now let us consider the value of this in the limit of large $\ell$, $\ell\gg c$.
First, let us expand $\frac{1}{Z}$ using the first order expansion $(1+x)^{-1}\approx 1-x$ for $|x|\ll 1$.
\begin{align}
    \frac{1}{Z} &= \frac{1}{\exp{\left(c+\ell\right)} + (K-1)\exp{\left(c\right)}}\\
    &=\frac{1}{\exp{\left(\ell+c\right)}} \frac{1}{1+ (K-1)\exp{\left(-\ell\right)}}\\
    &={\exp{\left(-\ell-c\right)}}\left(1- (K-1)\exp{\left(-\ell\right)} + O(\exp{\left(-\ell\right)}^2)\right).\label{eq:z_sm_expand}
\end{align}
Second, let us expand $\log Z$ using the first order expansion $\log(1+x)\approx x$ for $|x|\ll 1$.
\begin{align}
    \log Z &=\log\left(\exp{\left(c+\ell\right)} + (K-1)\exp{\left(c\right)}\right)\\
    &=\log\left(\exp{\left(c+\ell\right)}(1 + (K-1)\exp{\left(-\ell\right)}\right)\\
    &=c + \ell + \log\left(1 + (K-1)\exp{\left(-\ell\right)}\right)\\
    &=c + \ell + (K-1)\exp{\left(-\ell\right)}+O\left(\exp{\left(-\ell\right)}^2\right) \label{eq:logz_sm_expand}
    \end{align}

Taking Eqs~(\ref{eq:z_sm_expand},\ref{eq:logz_sm_expand}) and subbing back into Eq~\eqref{eq:h_form_vq}, keeping terms to first order in $\exp{\left(-\ell\right)}$ we get
\begin{equation}
    \mathcal{H}_\mathrm{softmax} \approx (K-1)\exp{\left(-\ell\right)}\left(1+\ell\right)
    \label{eq:sm_approx_h}
\end{equation}
as required.

\end{proof}

\newpage
\subsection{Experimental Evaluation}
\label{app:entropy}

Just as a check on these bounds, we calculate the entropy exactly and using these first-order approximations for both methods' worst-case scenarios.
We find the approximation to be highly accurate for inputs $>10$, with proportional error $\approx 10^{-6}$ for each.
\begin{figure}[h!]
    \centering
    \includegraphics[width=0.35\textwidth]{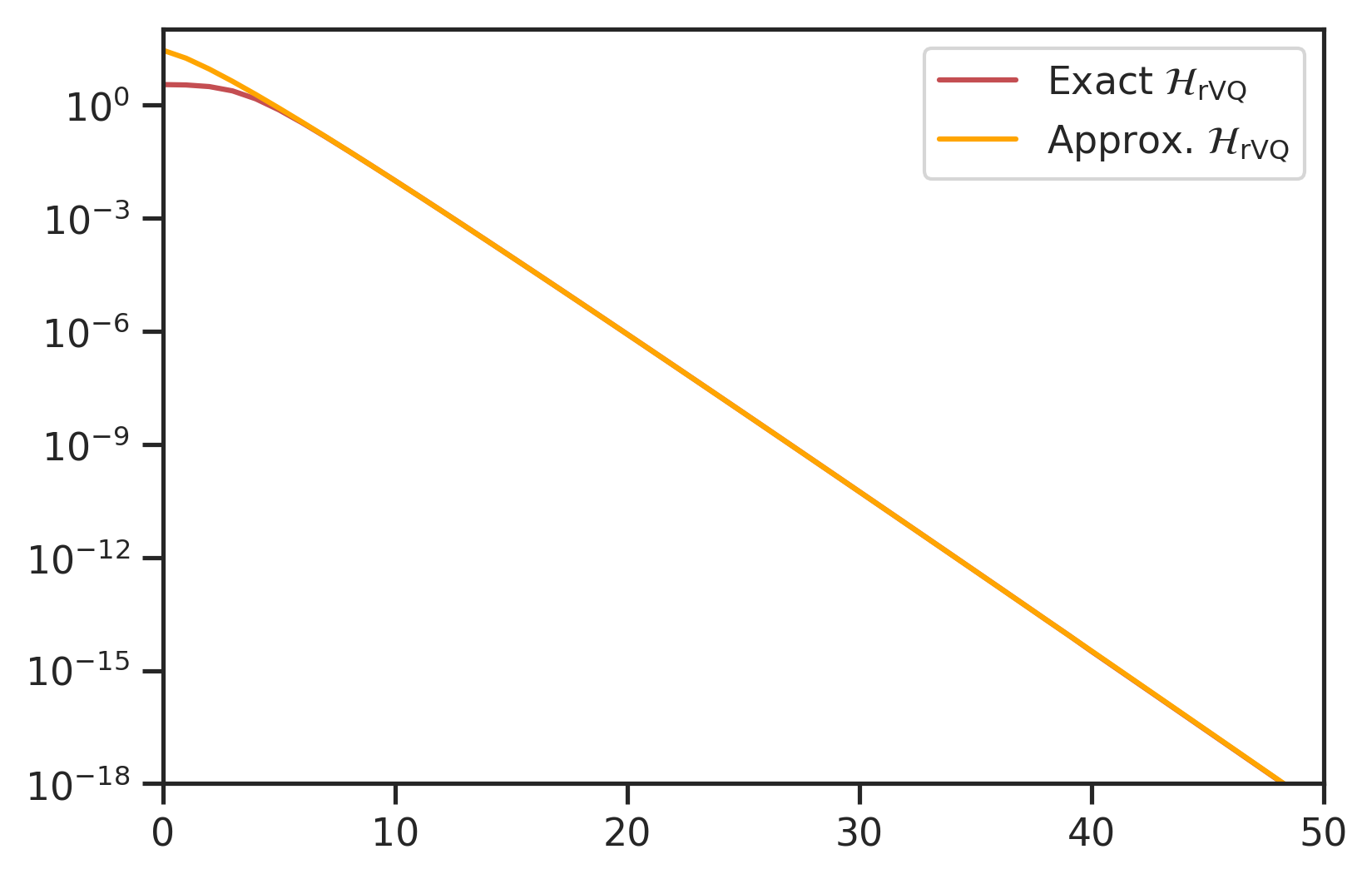}
    \caption{rVQ worst-case entropy as a function of $d$, calculated exactly and using Eq~\eqref{eq:rvq_approx_h}, for $\delta=1$. Note this is a logarithmic plot.}
    \label{fig:rvq_h}
\end{figure}

\begin{figure}[h!]
    \centering
    \includegraphics[width=0.35\textwidth]{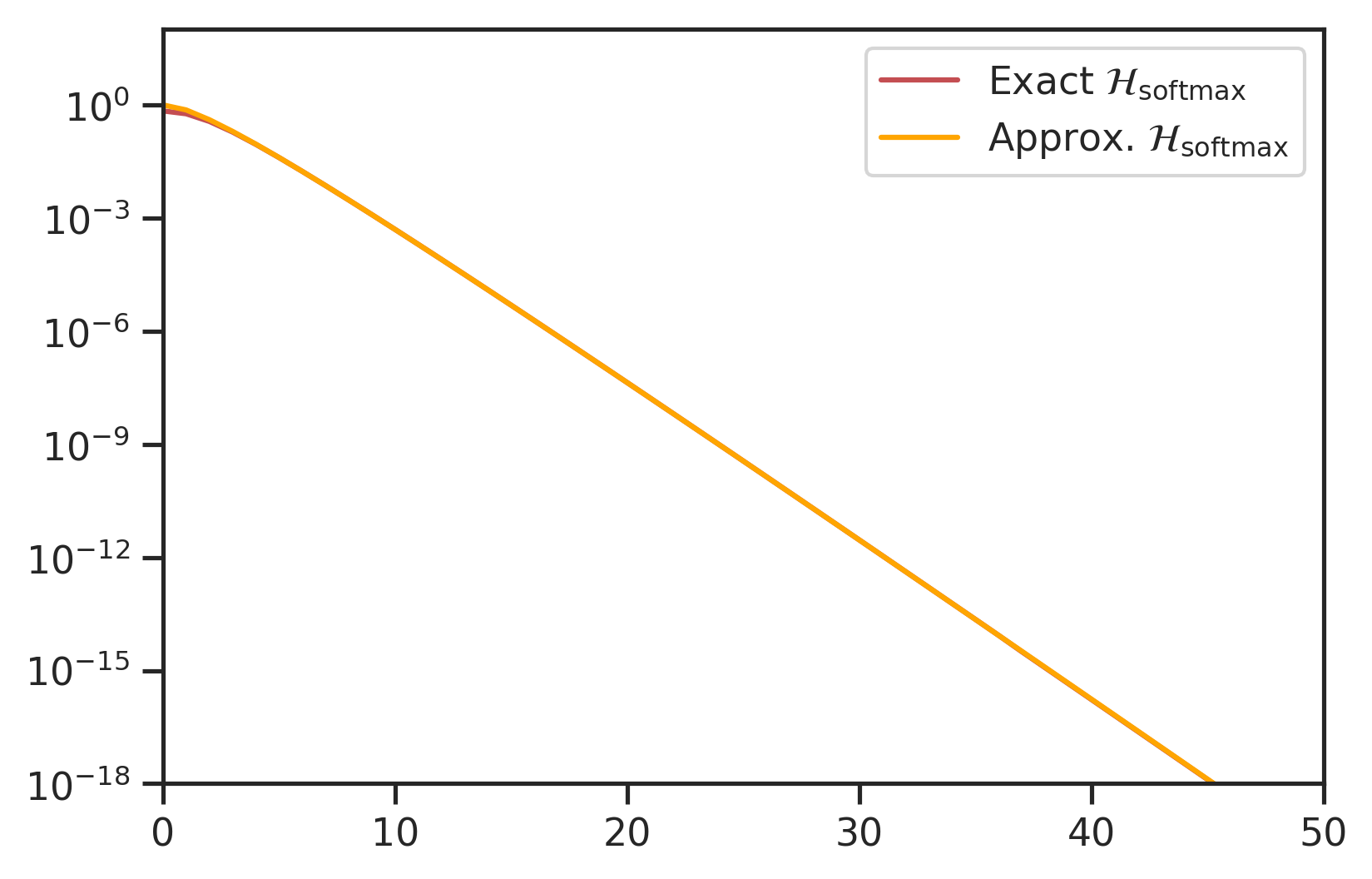}
    \caption{Softmax worst-case entropy as a function of $d$, calculated exactly and using Eq~\eqref{eq:sm_approx_h}, for $c=0$. Note this is a logarithmic plot.}
    \label{fig:sm_h}
\end{figure}

\begin{figure}[h!]
    \centering
    \includegraphics[width=0.35\textwidth]{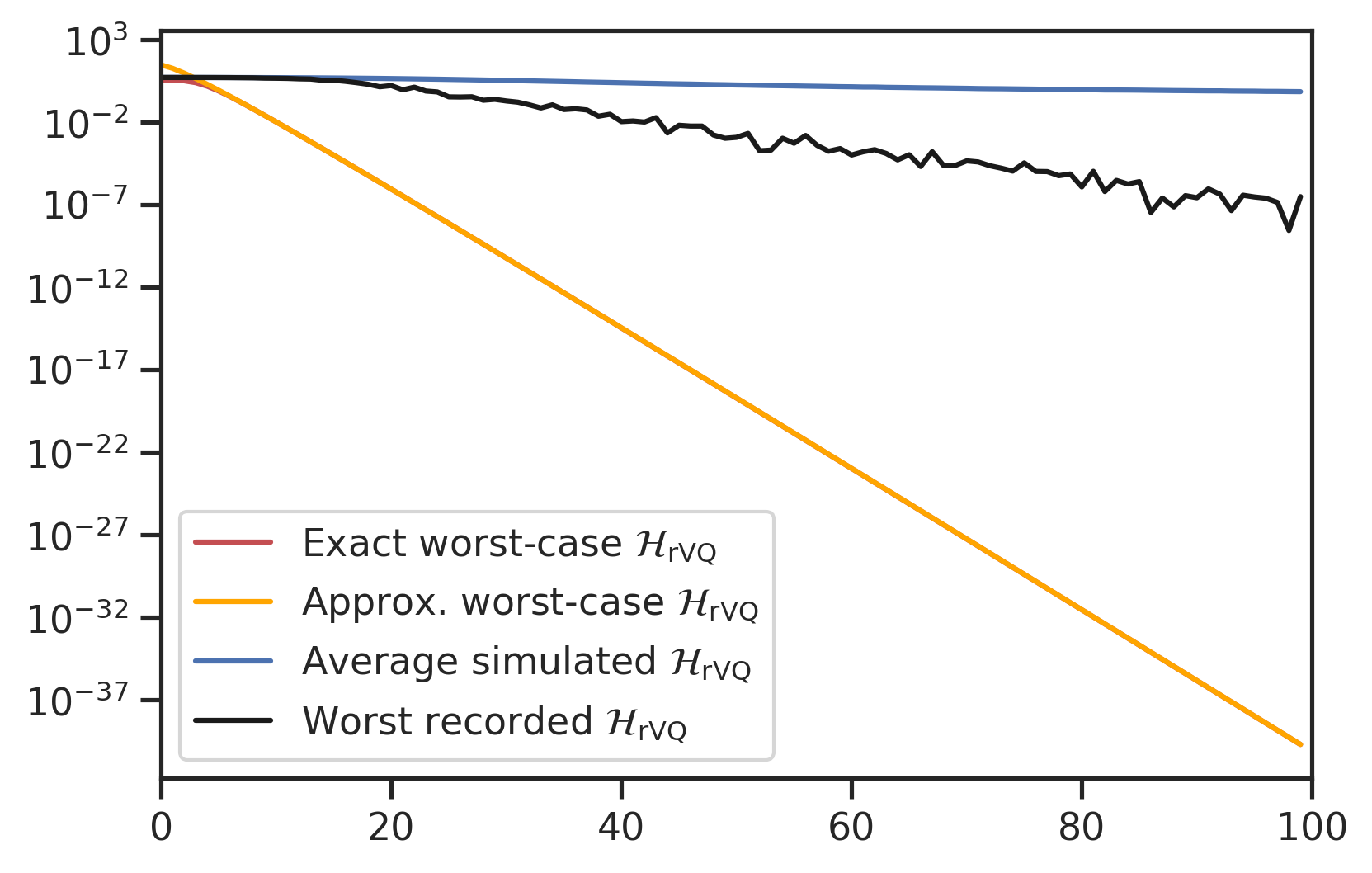}
    \caption{rVQ entropy as a function of $d$, calculated for the worst case both exactly and using Eq~\eqref{eq:rvq_approx_h}, for $\delta=1$, along with the average entropy from simulated codebooks with codebook embeddings uniform over the radius $0.5$ hypersphere and the worst recorded entropy from that simulation procedure at each distance. Note this is a logarithmic plot.}
    \label{fig:vq_h_sim}
\end{figure}

As a further check on the rVQ results in Fig~\ref{fig:vq_h_sim} we create random codebooks of embeddings uniformly distributed over the hypersphere with radius $0.5$ and calculate $\ent$ as a function of $d$.
We do this for 20,000 sampled codebooks per value of $d$, each of 256 entries,  in an embedding space with $d_e=32$.
The entropy we get from simulation is on a different trend entirely from the `worst-case' calculations, which makes sense as that worst-possible arrangement is a vanishingly unlikely to occur.

\newpage
\section{Samples and Reconstructions}

  \begin{figure}[h!]
  \centering
    \begin{subfigure}{0.5\linewidth}
      \hspace{1.5em}
      \makebox[0pt]{\rotatebox[origin=c]{90}{
        (a) C-10
      }\hspace*{2em}}%
    \begin{minipage}[c]{\textwidth}
      \begin{subfigure}{\linewidth - 2em}
      \adjincludegraphics[width=\linewidth,trim={0 {.872\height} 0 0},clip]{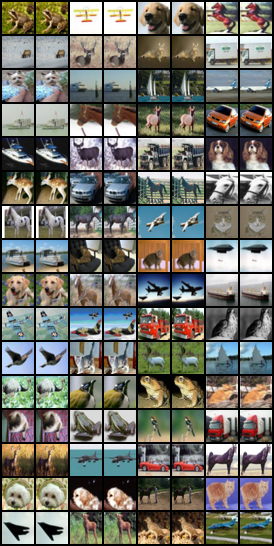}
    \vspace{-1.2em}
      \end{subfigure}\hfill
    \end{minipage}
    \end{subfigure}
    \begin{subfigure}{0.5\linewidth}
     \hspace{1.5em}
      \makebox[0pt]{\rotatebox[origin=c]{90}{
        (b) SVHN
      }\hspace*{2em}}%
    \begin{minipage}[c]{\textwidth}
      \begin{subfigure}{\linewidth - 2em}
\adjincludegraphics[width=\linewidth,trim={0 0 0 {.872\height}},clip]{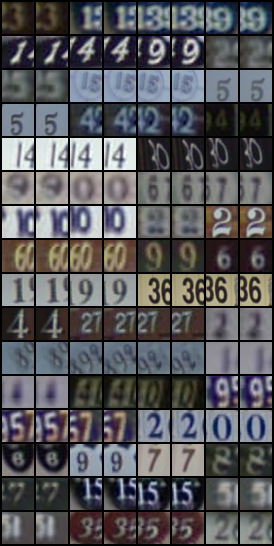}      \end{subfigure}\hfill
        \end{minipage}
    \end{subfigure}
    \begin{subfigure}{0.5\linewidth}
    \hspace{1.5em}
      \makebox[0pt]{\rotatebox[origin=c]{90}{
        (c) CelebA
      }\hspace*{2em}}%
      \begin{minipage}[c]{\textwidth}
      \begin{subfigure}{\linewidth - 2em}
\adjincludegraphics[width=\linewidth,trim={0 0 0 {.831\height}},clip]{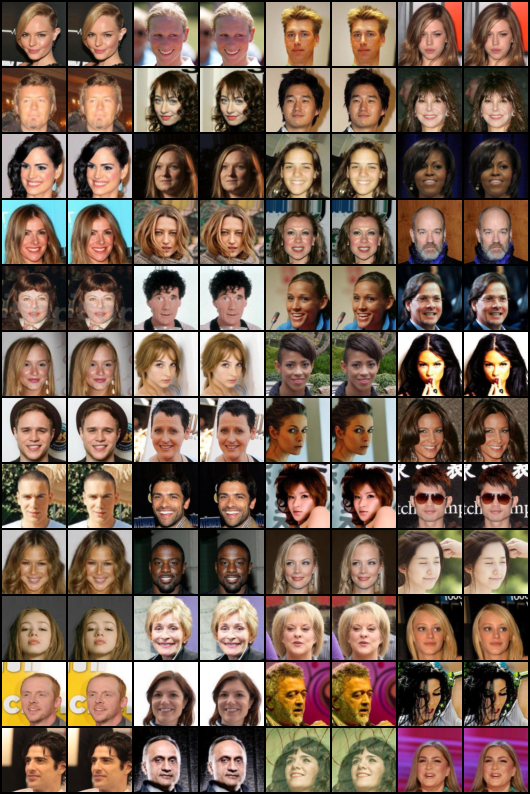}
    \end{subfigure}\hfill
    \end{minipage}
    \end{subfigure}
    \par\vskip \abovecaptionskip
        \caption{Reconstructions: We demonstrate our approach provides high quality reconstructions, for CIFAR-10, SVHN and CelebA. In each pair, left is the reconstruction, right the original.}
\label{fig:recons}
    \vspace{-1em}
  \end{figure}

\begin{figure}[h!]
\centering
\begin{subfigure}{0.5\linewidth}
  \hspace{1.5em}
  \makebox[0pt]{\rotatebox[origin=c]{90}{
    (a) C-10
  }\hspace*{2em}}%
\begin{minipage}[c]{\textwidth}
  \begin{subfigure}{\linewidth - 2em}
    \includegraphics[width=\linewidth]{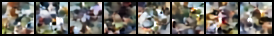}
  \end{subfigure}\hfill
  \begin{subfigure}{\linewidth - 2em}
  \adjincludegraphics[width=\linewidth,trim={0 0 0 {.744\height}},clip]{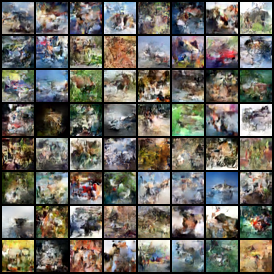}
    \vspace{-2.25em}
  \end{subfigure}\hfill
\end{minipage}
\end{subfigure}
\par\bigskip
\begin{subfigure}{0.5\linewidth}
 \hspace{1.5em}
  \makebox[0pt]{\rotatebox[origin=c]{90}{
    (b) SVHN
  }\hspace*{2em}}%
\begin{minipage}[c]{\textwidth}
  \begin{subfigure}{\linewidth - 2em}
    \includegraphics[width=\linewidth]{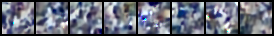}
  \end{subfigure}\hfill
  \begin{subfigure}{\linewidth - 2em}
  \adjincludegraphics[width=\linewidth,trim={0 0 0 {.744\height}},clip]{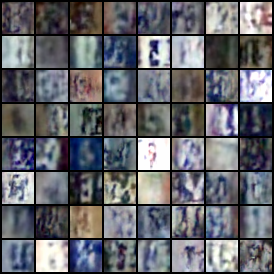}
\vspace{-1em}
  \end{subfigure}\hfill
    \end{minipage}
\end{subfigure}
\begin{subfigure}{0.5\linewidth}
\hspace{1.5em}
  \makebox[0pt]{\rotatebox[origin=c]{90}{
    (c) CelebA
  }\hspace*{2em}}%
  \begin{minipage}[c]{\textwidth}
    \begin{subfigure}{\linewidth - 2em}
    \includegraphics[width=\linewidth]{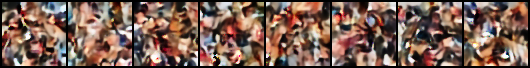}
  \end{subfigure}\hfill
  \begin{subfigure}{\linewidth - 2em}
\adjincludegraphics[width=\linewidth,trim={0 0 0 {.747\height}},clip]{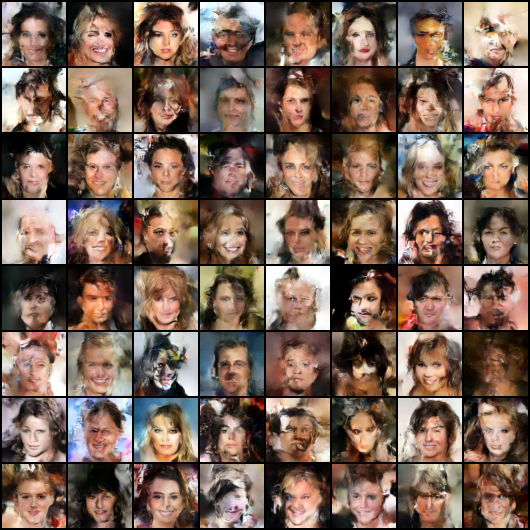}
  \end{subfigure}\hfill
\end{minipage}
\end{subfigure}
    \vspace{-1em}
\par\vskip\abovecaptionskip
    \caption{Sampling: we perform ancestral sampling for single-layer rVQ--VAE baselines (top row) and our $L=32$ models (middle and bottom), for CIFAR-10, SVHN and CelebA.}
    \vspace{-1em}
\label{fig:samples}
\end{figure}

\newpage
\section{Compression using RRVQ models}
\label{app:compress}
For our $L=5$ models, our latents $\vv{z}$ are in 5 layers of size $\v{M}=\{16\times16,8\times8,4\times4,2\times2,1\times1\}$.
$\{\v{E}_{\mu,\ell},\v{E}_{\Sigma,\ell}\}$.
For CIFAR-10 and SVHN these each containing $K=256$ codebook values $\in\mathbb{R}^{d_e}$, $d_e=128$, per layer.
For CelebA, we taper the number of embeddings per layer so $\v{K}=\{128, 64, 32, 16, 8\}$, $d_e=32$, and have networks layer-to-layer with fewer channels, for reasons of compute capacity.

\begin{figure}[h!]
    \centering
\includegraphics[width=0.4\linewidth]{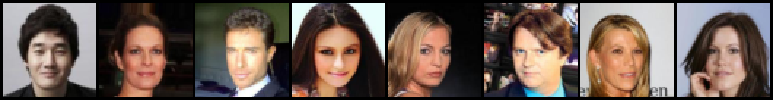}\\
\includegraphics[width=0.4\linewidth]{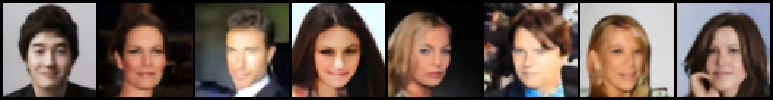}\\
\includegraphics[width=0.4\linewidth]{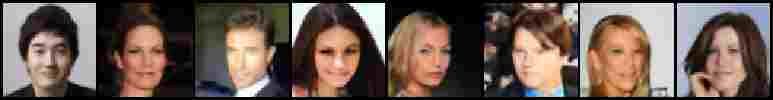}
    \caption{\textit{Top}: Original, \textit{Middle}: RRVQ $L=5$ compression, \textit{Bottom}: JPEG at same compression ratio. Best viewed zoomed in.}
    \label{fig:compression}
\end{figure}

In Fig~\ref{fig:compression} we compress (top) CelebA images using (middle) our $L=5$ model and (bottom) using JPEG to the same compression ratio (CR) [same experimental protocol as \citet{Gregor2016}.
We are compressing $64\times64$ images into $2275$ bits, a CR of $\frac{98304}{2275}\approx 43$.
Our approach outperforms JPEG, maintaining more visual information.
Unlike JPEG, ours does not introduce blocky artefacts.

\section{MLP rVQ-VAEs}
\begin{figure}[h!]
    \centering
    \adjincludegraphics[width=0.4\linewidth, trim=758 0 0 14, clip=true]{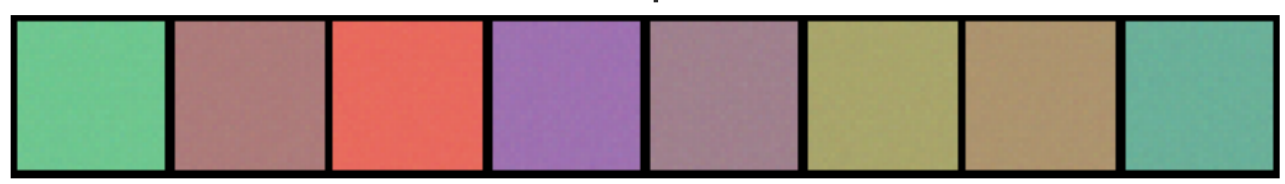}
    \caption{MLP-rVQ-VAE samples, trained on toy colour-swatch dateset.}
    \label{fig:mlp}
\end{figure}

For completeness' sake in Fig~\ref{fig:mlp} we train an MLP rVQ-VAE on our colour swatch data, to demonstrate that samples from such a model show consistent colour cast (further, samples show new colours beyond the training set).
That is, ancestral samples look like the training data (ie with consistent colour) unlike single-latent-layer convolutional models.

\label{app:mlp}
\end{document}